\documentclass[11pt]{article}
\usepackage[final]{acl}
\usepackage{times}
\usepackage{latexsym}
\usepackage[T1]{fontenc}
\usepackage[utf8]{inputenc}
\usepackage{microtype}
\usepackage{inconsolata}
\usepackage{graphicx}
\usepackage{url}
\usepackage{booktabs}
\usepackage{xspace}
\usepackage{amssymb}
\usepackage[table]{xcolor}
\usepackage{tikz}
\usetikzlibrary{arrows.meta}
\usetikzlibrary{calc}
\usepackage{pgfplots}
\pgfplotsset{compat=1.18}
\usepgfplotslibrary{fillbetween}
\usepgfplotslibrary{groupplots}
\usepackage{multirow}
\usepackage{makecell}
\usepackage{placeins}

\definecolor{baselineColor}{HTML}{4C92D9}
\definecolor{baselineColorv2}{HTML}{2B6CB0}
\definecolor{ultraColor}{HTML}{55C58F}
\definecolor{tickgreen}{HTML}{2E8B57}
\definecolor{crossred}{HTML}{C0392B}

\usepackage[most,skins,theorems]{tcolorbox}
\tcbset{
  aibox/.style={
    width=\linewidth,
    top=3pt,
    bottom=3pt,
    colback=blue!6!white,
    colframe=black,
    colbacktitle=black,
    coltitle=white,
    enhanced,
    sharp corners,
    boxrule=0.6pt,
    before skip=0cm,
    after skip=0cm,
    boxsep=2pt,
    left=4pt,
    right=4pt,
    fonttitle=\bfseries\small,
    fontupper=\small,
    halign title=flush left,
    before upper={\setlength{\parindent}{0pt}},
  }
}
\newtcolorbox{AIbox}[2][]{aibox,title=#2,#1}

\title{A Recipe for Long-Context Reasoning in Large Language Models \\
via On-Policy Optimization and Distillation}

\author{
\textbf{Miguel Moura Ramos}$^{1,2}$ \quad
\textbf{Duarte M. Alves}$^{1,2}$ \quad
\textbf{André F. T. Martins}$^{1,2,3,4}$ \quad    
\\
$^1$Instituto Superior Técnico, Universidade de Lisboa
\\
$^2$Instituto de Telecomunicações\quad 
$^3$TransPerfect\quad
$^4$ELLIS Unit Lisbon
}

\newcommand{\infiniteBench}{$\infty$Bench\xspace}
\newcommand{\infiniteBenchNS}{$\infty$Bench}
\newcommand{\longblocks}{\textsc{LongBlocks}\xspace}
\begin{document}

\maketitle

\begin{abstract}
Existing approaches to post-train models for long-context tasks face complementary limitations:
(i) supervised fine-tuning (SFT) provides stable supervision but suffers from exposure bias;
(ii) reinforcement learning methods such as Group Relative Policy Optimization (GRPO) train on model-generated trajectories but struggle with long-horizon credit assignment and sparse rewards;
and (iii) on-policy distillation (OPD) provides dense token-level guidance but does not directly optimize task rewards.
We study these complementary strategies for long-context alignment and derive a recipe that combines GRPO with OPD-style teacher guidance: the student learns from its own rollouts using outcome-level rewards, while a stronger teacher provides dense token-level regularization in place of the standard reference policy.
This is especially useful when process-level supervision is difficult to obtain.
To support this study, we introduce \longblocks, a synthetic multilingual dataset spanning multi-hop reasoning, contextual grounding, and long-form generation.
Through controlled ablations, we isolate the roles of cold-start initialization, teacher anchoring, and data mixing, showing that our recipe yields a more stable and effective path to long-context reasoning than GRPO or OPD while preserving short-context capabilities.\footnote{All resources are available on \href{https://huggingface.co/datasets/utter-project/LongBlocks}{Hugging Face}.}
\end{abstract}

\section{Introduction}

\begin{figure*}[!t]
\centering
\includegraphics[width=\linewidth]{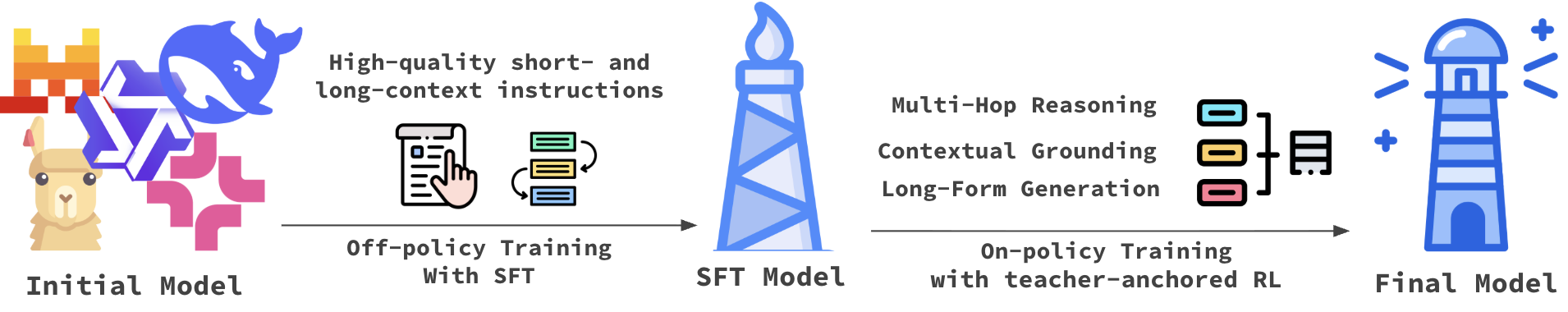}
\caption{Illustration of our post-training recipe for building a long-context reasoning model.}
\label{fig:example}
\end{figure*}

Real-world applications increasingly require large language models (LLMs) to reason over and integrate information from extremely long contexts. This arises in tasks requiring understanding large codebases \citep{jimenez2024swebench}, synthesizing information across collections of documents \citep{yang-etal-2018-hotpotqa}, and maintaining coherence across multi-session interactions spanning hundreds of thousands of tokens \citep{mei2025surveycontextengineeringlarge,liu-etal-2024-lost}.
Yet, aligning LLMs for long-context reasoning remains challenging due to multiple factors, including positional extrapolation, attention dilution, and exposure bias \citep{peng2024yarn,liu-etal-2024-lost,bengio2015scheduledsampling}.
These issues make long-context post-training qualitatively different from simply extending a short-context alignment recipe to longer sequences.
Existing post-training strategies lack a standardized, open recipe that reliably delivers strong long-context performance \citep{openai_gpt5_2025,yang2025qwen3technicalreport,2025arXiv250706261C}.
Current efforts are further constrained by limited data and compute: off-policy alignment datasets are typically small and focused on relatively simple retrieval and summarization tasks, which restricts the diversity of long-context supervision \citep{bai-etal-2024-longalign,chen2024longloraefficientfinetuninglongcontext}.
At the same time, long-context settings remain relatively underexplored for on-policy alignment \citep{ouyang2022training} and distillation \citep{agarwal2024onpolicy}. 
Existing long-context work has largely focused on preference optimization \citep{bai2024longwriterunleashing10000word,zhang-etal-2025-longreward} or on-policy methods with limited sampling \citep{wan2025qwenlongl1longcontextlargereasoning}, leaving the interaction between outcome-level rewards and dense teacher guidance poorly understood in long-context regimes.

Motivated by these gaps, we ask: \textbf{how can post-training components be integrated for efficient, stable, and generalizable long-context reasoning?}
We propose a two-stage recipe: an off-policy warm-up stage followed by teacher-anchored on-policy RL, where outcome-reward optimization on student-generated rollouts is performed jointly with OPD-style teacher guidance.
Warm-up improves rollout quality before RL, reward optimization adapts the model under its own trajectory distribution, and teacher guidance replaces the standard reference-policy anchor with dense token-level regularization from a stronger model.
This teacher signal provides intermediate supervision when process-level rewards are difficult to obtain, adapting GRPO to the long-context regime where sparse rewards, long rollouts, and short-context retention must be balanced.

To evaluate this recipe, we conduct a controlled empirical comparison of long-context alignment strategies in a Qwen3-based setup (Figure~\ref{fig:example}).
We isolate the roles of off-policy warm-up, outcome-reward GRPO, reference-policy anchoring, on-policy distillation, and teacher-anchored reward optimization under matched training budgets.
This allows us to measure how dense teacher guidance interacts with sparse rewards, long rollouts, and short-context retention.
To facilitate research on long-context post-training, we also introduce \textbf{\longblocks}, a synthetic multilingual long-context dataset comprising multi-hop reasoning, contextual grounding, and long-form generation examples.
Together, the recipe and dataset substantially improve long-context performance in a 1.7B model without sacrificing short-context performance on average.

\section{Preliminary}

\subsection{Language Generation as a MDP}
\label{subsec:token_mdp}
We formulate autoregressive language generation as a token-level Markov Decision Process (MDP)~\mbox{$\mathcal{M}=(S,A,r,T)$}, where $S$ is the set of token-prefix states, $A=\mathcal{V}$ is the action space given by the model vocabulary, $r$ is the reward function, and $T$ is the transition function.
At each generation step $t$, the state $s_t \in S$ consists of the prompt $p$ concatenated with all previously generated tokens: $s_t = [p_1, \ldots, p_M, o_1, \ldots, o_{t-1}]$.
The action at time $t$ is the next token $o_t \in A$.
The transition function is deterministic: after action $o_t$, the next state is formed by appending the token, $s_{t+1} = T(s_t,o_t) = [s_t; o_t]$.
The initial state $s_1$ is induced by the prompt $p \sim \mathcal{P}$, where $\mathcal{P}$ denotes the distribution over prompts.
An episode ends when the policy outputs an end-of-sequence token $[\text{eos}]$ or when the total token budget is reached.
A policy is an autoregressive conditional distribution $\pi_\theta(\cdot|s_t)$, and the reward for a trajectory $\tau = (p,o)$ is the sum of per-step rewards: $R(p,o)=\sum_{t=1}^{|o|} r(s_t,o_t)$.
We consider the undiscounted finite-horizon setting, which is standard for sequence generation tasks.
The agent typically maximizes a KL-regularized expected return~\citep{schulman2018equivalencepolicygradientssoft}:
\begin{align}
\mathcal{J}&(\pi_{\theta}) = 
\mathbb{E}_{p \sim \mathcal{P},\; o \sim \pi_{\theta}(\cdot|p)} \Bigg[ R(p,o) \notag \\
&\quad - \beta_{\mathrm{ref}} \sum_{t=1}^{|o|} D_{KL}\big(\pi_{\theta}(\cdot|s_t) \,\|\, \pi_{\mathrm{ref}}(\cdot|s_t)\big) 
\Bigg].
\end{align}
Here $\pi_{\mathrm{ref}}$ is a reference policy, and $\beta_{\mathrm{ref}}$ is a KL regularization weight used in RLHF to prevent the tuned policy from drifting into regions where the learned reward model is unreliable.
The KL term can be estimated on sampled trajectories using per-token log-ratios, $\log \pi_{\theta}(o_t|s_t) - \log \pi_{\mathrm{ref}}(o_t|s_t)$.
Following prior work~\citep{liu2025understanding, yu2025dapo, olmo2025olmo3, k2team2025k2v2360openreasoningenhancedllm}, we set $\beta_{\mathrm{ref}} = 0$, eliminating the need for a reference policy and reducing memory and compute costs.

\subsection{Off-Policy Training}
\label{subsec:off_policy}
Off-policy supervision, primarily via supervised fine-tuning (SFT), provides a stable and data-efficient approach to bootstrap long-context behavior and can be viewed as a form of imitation learning \citep{pmlr-v15-ross11a} for language models.
Formally, given an expert dataset $\mathcal{D}$ of input--output pairs $(x,y)$, we optimize the model parameters $\theta$ by minimizing the token-level negative log-likelihood:
\begin{equation}
    \mathcal{L}_{\text{SFT}}(\theta) = \mathbb{E}_{(x,y) \sim \mathcal{D}} \left[ -\sum_{t=1}^{|y|} \log \pi_{\theta}(y_t \mid x, y_{<t}) \right].
\end{equation}
This objective provides dense token-level supervision, making it effective for instilling task adherence \citep{raffael2020,ouyang2022training}.
When a strong teacher model is available, knowledge distillation \citep{hinton2015distillingknowledgeneuralnetwork} can improve training by aligning the student policy $\pi_\theta$ with the teacher policy $\pi_{\text{T}}$.
This is typically done by minimizing the token-level Kullback--Leibler divergence between the two distributions:
\begin{align}
    \mathcal{L}&_{\text{KD}}(\theta) = \mathbb{E}_{(x,y) \sim \mathcal{D}} \Big[ \notag \\
    &\quad \sum_{t=1}^{|y|} D_{KL}\big(\pi_{\text{T}}(\cdot \mid x, y_{<t}) \,\|\, \pi_{\theta}(\cdot \mid x, y_{<t})\big) \Big].
\end{align}
This distillation objective encourages the student to match the teacher's token distributions at each prefix, providing fine-grained guidance that can improve generalization beyond standard SFT.
However, because both SFT and distillation rely on static expert trajectories, they remain off-policy and suffer from exposure bias \citep{ranzato2016sequenceleveltrainingrecurrent,bengio2015scheduledsampling}: they optimize only on data-distribution prefixes and do not train recovery from the model's own off-distribution states.
This can degrade generation quality by amplifying errors during inference \citep{arora-etal-2022-exposure,zhang-etal-2019-bridging,chiang-chen-2021-relating}, especially in long-context settings, where retrieval distractors, positional extrapolation, length calibration, and optimization instability can compound over long rollouts.

\subsection{On-Policy Training}
\label{subsec:on_policy}

On-policy RL approaches, such as Group Relative Policy Optimization (GRPO) \citep{shao2024deepseekmathpushinglimitsmathematical}, address an important limitation of off-policy training approaches by conditioning updates on the model’s own rollouts.
GRPO avoids exposure bias by optimizing the policy under the distribution induced by its own rollouts, enabling adaptation to long-range dependencies. 
Unlike standard PPO \citep{ppo}, GRPO simplifies value estimation by removing the need for a separate value function critic. 
Instead, it estimates the baseline from a group of sampled completions $\{o_1, o_2, \ldots, o_G\}$ for a given prompt $p$. The objective function, with $\beta_{\mathrm{ref}}=0$, can be defined as:
\begin{align}
\mathcal{J}_{\text{GRPO}}(\theta)
&=
\mathbb{E}_{p \sim \mathcal{P},\; \{o_i\}_{i=1}^G \sim \pi_{\theta_{\text{old}}}(\cdot|p)}
\Bigg[
\frac{1}{G} \sum_{i=1}^G \nonumber\\ \frac{1}{|o_i|} \sum_{t=1}^{|o_i|}
\Bigg.
&\Bigg.
\min\Big(
\rho_{i,t}(\theta)\,\hat{A}_{i,t},\;
\bar{\rho}_{i,t}(\theta)\,\hat{A}_{i,t}
\Big)
\Bigg],
\label{eq:GRPO-obj}
\end{align}
where
\[
\rho_{i,t}(\theta)
=
\frac{\pi_\theta(o_{i,t} \mid p, o_{i,<t})}{\pi_{\theta_{\text{old}}}(o_{i,t} \mid p, o_{i,<t})},
\]
\[
\bar{\rho}_{i,t}(\theta)
=
\mathrm{clip}\!\left(\rho_{i,t}(\theta),\, 1-\epsilon,\, 1+\epsilon\right),
\]
and
{\small\[
\hat{A}_{i,t}
=
\frac{R(p, o_i)-\mathrm{mean}\!\left( \{ R(p, o_1), \dots, R(p, o_G) \} \right)}
{\mathrm{std}\!\left( \{ R(p, o_1), \dots, R(p, o_G) \} \right)}.
\]}

Standalone GRPO is attractive for long-context alignment because it trains on model-generated trajectories, but its sparse rewards make optimization sample-inefficient and high-variance.
It also remains susceptible to reward hacking \citep{liu2025understanding,yu2025dapo}, especially during long-context training where trajectory-level rewards are delayed and the lack of fine-grained supervision can hurt stability.
An alternative approach that provides denser supervision is on-policy distillation (OPD), which guides the student policy toward a strong teacher by minimizing the reverse token-level KL divergence on student-generated trajectories, thereby providing token-level guidance while remaining on-policy \citep{agarwal2024onpolicy,yang2025qwen3technicalreport,lu2025onpolicydistillation}.
Unlike regularization toward a fixed reference policy, OPD aligns the student to the teacher on prefixes induced by student rollouts.
Using reverse KL makes the distillation objective student-centered, focusing supervision on trajectories the student visits, which makes it a natural fit for on-policy RL.
We define the OPD objective as
\begin{align}
\mathcal{L}_{\text{OPD}}&(\theta)
=
\mathbb{E}_{p \sim \mathcal{P},\; o \sim \pi_\theta(\cdot|p)}
\left[
\frac{1}{|o|}\sum_{t=1}^{|o|}
\right.
\nonumber\\
&\left.
D_{KL}\big(
\pi_\theta(\cdot \mid p, o_{<t})
\,\|\, 
\pi_{\text{T}}(\cdot \mid p, o_{<t})
\big)
\right].
\label{eq:opd}
\end{align}
However, OPD alone does not perform reward optimization and, therefore, cannot fully exploit reward-based feedback.
In Section~\ref{sec:longcontext}, we build on this observation by pairing GRPO-style outcome optimization with an on-policy teacher anchor, yielding a teacher-anchored reward-optimization objective for long-context alignment.

\section{Recipe for Long-Context Reasoning}
\label{sec:longcontext}

The preceding discussion suggests that no single ingredient is sufficient: off-policy supervision is stable but static, sparse-reward RL is adaptive but unstable, and OPD provides dense guidance but does not optimize task rewards.
As shown in Figure~\ref{fig:example}, we therefore use a unified training pipeline that combines reward optimization with dense teacher guidance.
This enables more stable policy improvement at extended sequence lengths while preserving short-context performance on average.

\paragraph{Cold-start Stage: Off-policy warm-up for stable long-horizon rollouts.}
The pipeline begins with SFT warm-up, which provides dense token-level supervision, improves long-context rollout coherence, and stabilizes subsequent on-policy learning. This is especially important in long-horizon settings, where low-quality or high-variance rollouts can make advantage estimation and KL-regularized updates brittle~\citep{stiennon2020learning,ouyang2022training,bai2022training}. 
We validate this stage in \S\ref{sec:experiments:ablations}, including comparisons between SFT- and KD-based initialization.

\paragraph{RL Stage: Teacher-anchored on-policy reward optimization.}
After initialization, we perform on-policy RL with GRPO (Eq.~\ref{eq:GRPO-obj}) to optimize long-horizon task rewards under the model's own rollout distribution.
In long-context settings, however, optimizing sparse outcome rewards alone can be unstable, degrading reasoning structure and short-context skills.
We therefore replace the standard reference-policy anchor in KL-regularized GRPO with a stronger teacher.
Although both use token-level KL regularization, they serve different roles: reference-policy KL is primarily a conservative constraint that limits drift from an earlier policy, whereas teacher KL provides a distillation signal toward a stronger policy.
Thus, rather than merely constraining the update, teacher anchoring supplies dense directional supervision on student-generated trajectories while preserving the outcome-reward objective.
Given a prompt \(p \sim \mathcal{P}\), we sample a group of \(G\) rollouts \(\{o_i\}_{i=1}^G \sim \pi_{\theta_{\text{old}}}(\cdot \mid p)\) and compute normalized advantages \(\hat A_{i,t}\).
Using the importance ratios \(\rho_{i,t}(\theta)\) and clipped ratios \(\bar{\rho}_{i,t}(\theta)\) from Eq.~\ref{eq:GRPO-obj}, we maximize
\begin{align}
&\mathcal{J}_{\text{ours}}(\theta)
=
\mathbb{E}_{p \sim \mathcal{P},\, \{o_i\}_{i=1}^G \sim \pi_{\theta_{\text{old}}}(\cdot \mid p)}
\Bigg[
\frac{1}{G}
\sum_{i=1}^G
\nonumber\\
&\quad
\frac{1}{|o_i|}
\sum_{t=1}^{|o_i|}
\Bigg(
\min\left(
\rho_{i,t}(\theta)\,\hat A_{i,t},\;
\bar{\rho}_{i,t}(\theta)\,\hat A_{i,t}
\right)
\nonumber\\
&\quad
-\;\beta\,
D_{\mathrm{KL}}\!\left(
\pi_\theta(\cdot \mid p, o_{i,<t})
\,\|\,
\pi_{\text{T}}(\cdot \mid p, o_{i,<t})
\right)
\Bigg)
\Bigg],
\label{eq:ours}
\end{align}
where \(\beta \ge 0\) controls the strength of teacher regularization.
The first term is the GRPO clipped surrogate for outcome-reward optimization, and the second is the teacher-anchoring term.

\begin{table*}[!htbp]
\centering
\small
\renewcommand{\arraystretch}{1.05}
\setlength{\tabcolsep}{6pt}
\begin{tabular}{l p{0.45\linewidth} r}
\toprule
\textbf{Source} & \textbf{Coverage} & \textbf{Samples} \\
\midrule
Institutional Books \citep{cargnelutti2025institutionalbooks10242b}
& 3 natural languages (de, en, fr)
& \textbf{107,817} \\
FineWeb2-HQ \citep{messmer2025multilingdatacomp}
& 9 natural languages (de, el, es, fr, hu, it, nl, sv, zh)
& \textbf{39,083} \\
ArXiv \citep{clement2019usearxivdataset}
& English
& \textbf{13,963} \\
Project Gutenberg Books \citep{ProjectGutenberg}
& 10 natural languages (de, en, fr, hu, it, nl, pl, pt, sv, zh)
& \textbf{10,987} \\
Wikipedia \citep{wikidump}
& 30 natural languages (ar, ca, da, de, el, es, \ldots, uk, zh)
& \textbf{10,527} \\
Stack-Edu \citep{allal2025smollm2smolgoesbig}
& 15 code/markup languages (Python, C++, \ldots, Markdown)
& \textbf{5,859} \\
StackExchange \citep{kandpal2025common_se}
& 8 natural languages (ar, en, hu, nl, ru, sv, tr, zh)
& \textbf{4,983} \\
\midrule
\textbf{Total} & & \textbf{193,219} \\
\bottomrule
\end{tabular}
\caption{Multilingual sources used for \longblocks.
Appendix~\ref{appendix:synthetic-qa} details the language-wise composition.}
\label{tab:dataset-sources}
\end{table*}

\paragraph{\longblocks{}: A Multilingual Dataset to Tailor LLMs for Long-Context Reasoning.}
To support long-context training, we introduce \longblocks{}, a multilingual synthetic dataset used for both off-policy warm-up and on-policy alignment.
It contains 193{,}219 question--answer pairs generated from the raw document sources summarized in Table~\ref{tab:dataset-sources}, covering more than 30 natural languages and 15 programming and markup languages.
We construct \longblocks{} with a two-stage pipeline: first, we sample long, self-contained documents and prompt \href{https://huggingface.co/Qwen/Qwen3-Next-80B-A3B-Thinking}{\texttt{Qwen3-Next-80B-A3B-Thinking}} with the templates in Appendix~\ref{appendix:synthetic-qa} to generate question--answer pairs requiring evidence integration across extended contexts; second, we apply fuzzy deduplication and LLM-as-a-judge verification to remove malformed, weakly grounded, or locally answerable examples.
Unlike prior long-context instruction datasets that primarily target document QA or instruction following, \longblocks{} is designed to provide reasoning-oriented long-context supervision, with tasks and answers that encourage evidence aggregation, multi-hop reasoning, summarization, classification, and grounded long-form generation.
We allocate examples uniformly across these task families at generation time, so that the mixture is not dominated by any single supervision format.
Appendix~\ref{appendix:synthetic-qa} compares \longblocks{} with existing long-context instruction datasets and details its benchmark-overlap controls and task-family balancing.
To mitigate catastrophic forgetting across context lengths, we maintain a fixed 10/90 token-level mixture of short- and long-context data during both SFT and RL, drawing short-context samples from \href{https://huggingface.co/datasets/nvidia/Nemotron-Post-Training-Dataset-v2}{\texttt{Nemotron-Post-Training-Dataset-v2}} and long-context samples from \longblocks{}.
We analyze this mixture in \S\ref{sec:experiments:ablations}, showing that the 10/90 ratio is sufficient to preserve average short-context performance, whereas larger short-context shares increasingly dilute long-context gains.

\paragraph{Training details.}
We initialize from \href{https://huggingface.co/Qwen/Qwen3-1.7B}{\texttt{Qwen3-1.7B}} and apply the two-stage pipeline in \S\ref{sec:longcontext}.
We extend the context length to 128K without changing the original positional parameterization, setting the RoPE base frequency to $1\times10^{6}$ \citep{su2021roformer}.
For SFT, we train for two epochs on the data mixture above with a learning rate of $1\times10^{-5}$ and a batch size of four million tokens.
We optimize target-token cross-entropy with Adam \citep{kingma2017adam}, weight decay $0.01$, \texttt{bfloat16} mixed precision, sequence packing \citep{raffael2020}, and document masking to prevent cross-document attention \citep{grattafiori2024llama3herdmodels}.
For RL, we optimize Eq.~\ref{eq:ours} on the same mixture.
Following \citet{olmo2025olmo3}, we use verifiable rewards for short-context tasks: SymPy for math, unit tests for coding, and binary constraint satisfaction for instruction following.
For chat and long-context tasks, we use \href{https://huggingface.co/Qwen/Qwen3-4B}{\texttt{Qwen3-4B}} as a reference-conditioned LM judge that assigns binary rewards by comparing responses to references, following \citet{lee2024rlaif}.
Appendix~\ref{appendix:synthetic-qa} validates the Qwen3-4B reward judge against 250 human annotations (84.8\% agreement) and against Qwen3-32B and Gemma-3-27B on 10,000 examples each (91.6\% and 87.0\% agreement, respectively).
We use outcome-level rewards because many tasks lack canonical step-by-step decompositions.
The teacher is \href{https://huggingface.co/Qwen/Qwen3-32B}{\texttt{Qwen3-32B}}, with $\beta=0.5$; we use a same-family teacher to enable exact token-level KL over a shared token vocabulary and analyze sensitivity in \S\ref{sec:experiments:ablations}.
RL uses Adam with a learning rate of $1\times10^{-6}$, $(\beta_1,\beta_2)=(0.9,0.95)$, gradient clipping $1.0$, constant learning rate, clipping threshold $0.2$, and $8$ rollouts per prompt.
Training used 4 NVIDIA H200 GPUs.
All RL comparisons use a one-epoch budget and identical rollout settings.

\section{Experiments}

\subsection{Experimental Setup}
\label{sec:experiments:setup}

We compare our final model against the base model, \texttt{Qwen3-1.7B}, and the SFT model from the cold-start stage as internal controls, together with representative 3B--4B long-context models as contextual reference points, including a YaRN-based context-extension model from the same family.

\paragraph{Decoding configuration.}
All models are evaluated with \href{https://github.com/EleutherAI/lm-evaluation-harness}{\texttt{lm-eval-harness}} using a common decoding configuration: temperature $0.6$, top-$p=0.95$, top-$k=20$, presence penalty $1.5$, and a maximum generation length of $32{,}768$ tokens, following \citet{yang2025qwen3technicalreport}. 
To reduce variance from stochastic decoding, we run each benchmark $5$ times and report the mean, with standard deviations provided in Appendix~\ref{sec:appendix-detailed-results}, as in \citet{olmo2025olmo3}.

\paragraph{Short-context evaluation.}
We evaluate on IFEval~\citep{zhou2023instructionfollowing}, IFBench~\citep{pyatkin2025generalizingverifiableinstructionfollowing}, MMLU-Pro~\citep{NEURIPS2024_ad236edc}, GPQA $\blacklozenge$~\citep{rein2024gpqa}, GSM8K~\citep{cobbe2021training}, MATH-500~\citep{lightman2023lets}, HumanEval~\citep{chen2021codex}, HumanEval+~\citep{evalplus}, MMMLU~\citep{openai_mmmlu_2024}, and MGSM~\citep{shi2022language}, covering instruction following, reasoning, multilinguality, and coding.

\paragraph{Long-context evaluation.}
We evaluate RULER~\citep{hsieh2024ruler} and OneRULER~\citep{kim2025one} at sequence lengths up to 128K, and LongBench~\citep{bai-etal-2024-longbench} and \infiniteBench~\citep{zhang-etal-2024-bench} using a maximum context length of 128K, covering retrieval, diagnostic tasks, multilingual aggregation, and long-context reasoning.

\subsection{Main Results and Findings}
\label{sec:experiments:results}

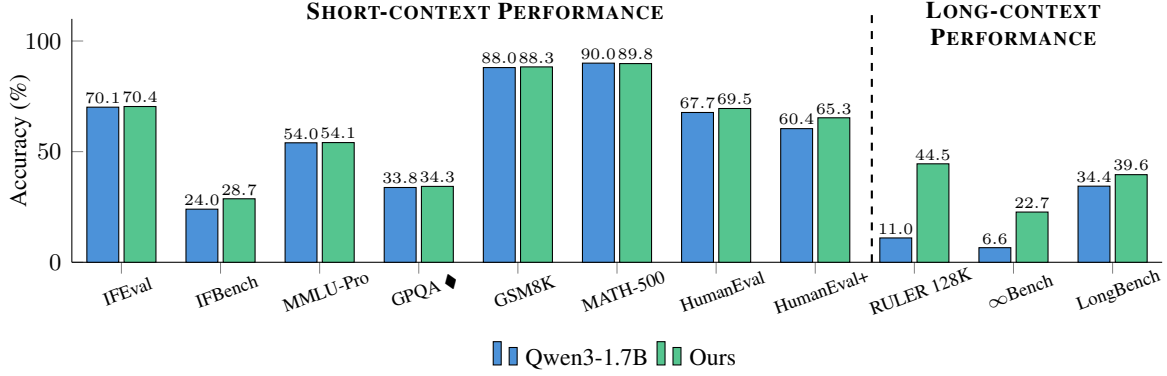
\begin{figure*}[!ht]
    \centering
    \begin{tikzpicture}[trim axis left, trim axis right]
        \begin{axis}[
            axis lines*=left, 
            ybar=2pt,
            bar width=12pt,
            width=\linewidth,
            height=5cm,
            enlarge x limits=0.05,
            ylabel={Accuracy (\%)},
            ylabel style={font=\small, yshift=-10pt},
            yticklabel style={font=\small, xshift=0pt},
            symbolic x coords={
                IFEval, IFBench, MMLU-Pro, GPQA $\blacklozenge$, GSM8K, MATH-500, HumanEval, HumanEval+,
                RULER 128K, $\infty$Bench, LongBench
            },
            xtick=data,
            xticklabel style={rotate=20, anchor=center, yshift=-7.0pt, font=\scriptsize},
            legend style={
                at={(0.5,-0.3)},
                anchor=north,
                legend columns=2,
                font=\small,
                column sep=1pt,
                draw=none
            },
            ymin=0, ymax=110,
            clip=false,      
            nodes near coords,
            every node near coord/.append style={
                text=black,
                font=\tiny,
                /pgf/number format/fixed,
                /pgf/number format/fixed zerofill,
                /pgf/number format/precision=1,
                anchor=center,
                yshift=3.5pt
            },
        ]

        \addplot+[fill=baselineColor, draw=black] coordinates {
            (IFEval, 70.1) (IFBench, 24.0) (MMLU-Pro, 54.0) (GPQA $\blacklozenge$, 33.8)
            (GSM8K, 88.0) (MATH-500, 90.0) (HumanEval, 67.7) (HumanEval+, 60.4)
            (RULER 128K, 11.0) ($\infty$Bench, 6.6) (LongBench, 34.4)
        };
    
        \addplot+[fill=ultraColor, draw=black] coordinates {
            (IFEval, 70.4) (IFBench, 28.7) (MMLU-Pro, 54.1) (GPQA $\blacklozenge$, 34.3)
            (GSM8K, 88.3) (MATH-500, 89.8) (HumanEval, 69.5) (HumanEval+, 65.3)
            (RULER 128K, 44.5) ($\infty$Bench, 22.7) (LongBench, 39.6)
        };

        \draw[dashed, thick] ([xshift=-16pt]axis cs:RULER 128K,0) -- ([xshift=-16pt]axis cs:RULER 128K,115);

        \node[anchor=south, font=\small\bfseries] at ([xshift=25pt]axis cs:GPQA $\blacklozenge$, 105) {\textsc{Short-context Performance}};
        \node[anchor=south, font=\small\bfseries] at (axis cs:$\infty$Bench, 90) {\makecell{\textsc{Long-context}\\\textsc{Performance}}};
        
        \legend{Qwen3-1.7B, Ours}
        \end{axis}
    \end{tikzpicture}
    \caption{Short- and long-context benchmark performance for the base model and the final recipe.}
    \label{fig:main_results}
\end{figure*}

\paragraph{Our recipe improves long-context performance without degrading short-context performance on average.}
Figure~\ref{fig:main_results} and Table~\ref{tab:ruler} show substantial long-context gains while maintaining overall short-context accuracy.
Most short-context benchmarks remain stable or improve, with the largest gains in code generation; minor negative fluctuations fall within the observed run-to-run variability (Appendix~\ref{sec:appendix-detailed-results}, which reports standard deviations across five runs). 
\begin{table}[!ht]
\centering
\small
\setlength{\tabcolsep}{4pt}
\begin{tabular}{@{}lccccccc@{}}
\toprule
\multirow{2}{*}{\textbf{Model}} & \multicolumn{7}{c}{\textbf{\textsc{RULER}}} \\ \cmidrule(l){2-8} 
 & \textbf{Avg.} & \textbf{4K} & \textbf{8K} & \textbf{16K} & \textbf{32K} & \textbf{64K} & \textbf{128K} \\ \midrule
Qwen3-1.7B  & 64.5 & 87.4 & 84.2 & 81.0 & 72.0 & 51.5 & 11.0 \\
Ours  & \textbf{74.8} & \textbf{88.3} & \textbf{85.4} & \textbf{84.7} & \textbf{80.4} & \textbf{65.6} & \textbf{44.5} \\
\bottomrule
\end{tabular}
\caption{RULER performance across sequence lengths.}
\label{tab:ruler}
\end{table}

On long-context benchmarks, average RULER performance rises from 64.5 to 74.8, with larger gains at longer sequence lengths (11.0$\rightarrow$44.5 at 128K).
The gains extend beyond retrieval-style diagnostics, including Multi-Document QA (23.4$\rightarrow$43.1) and Code Completion (20.3$\rightarrow$26.9) on LongBench, as well as English Multiple Choice (18.8$\rightarrow$30.1) and Math.Find (4.9$\rightarrow$10.6) on \infiniteBench.
However, the hardest long-horizon categories, including Code.Run and Math.Calc, remain an important frontier.
Appendix~\ref{sec:appendix-detailed-results} also provides the full per-task and multilingual breakdowns.

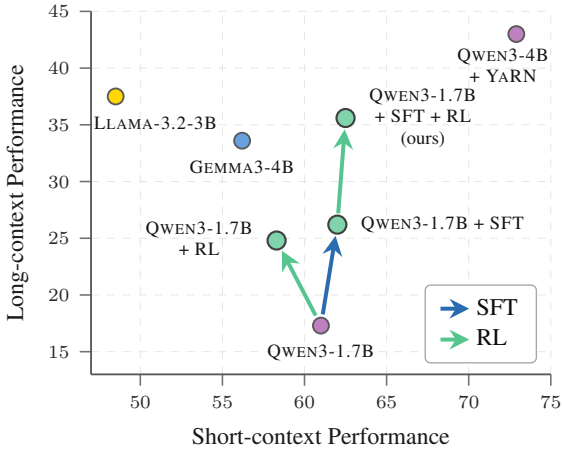
\begin{figure}[!t]
\centering
\tikzset{
  modeldot/.style   ={draw=black!50, fill=white, line width=0.6pt},
  green_arrow/.style ={draw=ultraColor, line width=1.5pt, -{Stealth[length=2.6mm,width=2.6mm]}},
  blue_arrow/.style ={draw=baselineColorv2, line width=1.5pt, -{Stealth[length=2.6mm,width=2.6mm]}},
  modeltext/.style  ={font=\scriptsize, text=black!92},
}

\begin{tikzpicture}
\begin{axis}[
  width=0.38\textwidth,
  height=0.38\textwidth,
  scale only axis,
  xmin=47, xmax=75,
  ymin=13, ymax=45,
  xtick={50,55,60,65,70,75},
  ytick={15,20,25,30,35,40,45},
  grid=major,
  major grid style={draw=black!10, line width=0.3pt, dashed},
  axis x line*=bottom,
  axis y line*=left,
  axis line style={draw=black!65, line width=0.7pt},
  tick style={draw=black!55, line width=0.6pt},
  tick align=outside,
  ticklabel style={font=\scriptsize, text=black!85},
  label style={font=\small, text=black!90},
  xlabel={Short-context Performance},
  ylabel={Long-context Performance},
  clip=false,
  legend style={
    draw=black!35, fill=white, font=\small, text=black!95,
    rounded corners=2pt, line width=0.5pt,
    inner xsep=5pt, inner ysep=3pt
  },
  legend cell align=left,
  legend pos=south east,
]

\newcommand{\modelpoint}[6]{
  \coordinate (#1) at (axis cs:#2,#3);
  \filldraw[fill=#6, draw=black!65, line width=0.7pt] (#1) circle (3.0pt);
  \node[modeltext,#5] at (#1) {#4};
}

\newcommand{\ourmodelpoint}[5]{
  \coordinate (#1) at (axis cs:#2,#3);
  \filldraw[fill=ultraColor!75!white, draw=black!75, line width=0.8pt] (#1) circle (3.4pt);
  \node[modeltext,#5] at (#1) {#4};
}

\modelpoint{qwen17}{61.0}{17.3}
  {\textsc{Qwen3-1.7B}}{anchor=center,yshift=-10pt}{violet!50}

\modelpoint{llama32}{48.5}{37.5}
  {\textsc{Llama-3.2-3B}}{anchor=center, xshift=15pt, yshift=-10pt}{yellow!75!orange}

\modelpoint{gemma4b}{56.2}{33.6}
  {\textsc{Gemma3-4B}}{anchor=center, yshift=-10pt}{baselineColor!85!white}

\modelpoint{qwen4b}{72.9}{43.0}
  {\makecell{\textsc{Qwen3-4B}\\\textsc{+ YaRN}}}{anchor=center, yshift=-12.5pt, xshift=-5pt}{violet!50}

\ourmodelpoint{sft17}{62.0}{26.2}
  {\textsc{Qwen3-1.7B + SFT}}{anchor=west, xshift=5pt}

\ourmodelpoint{ours17}{62.5}{35.6}
  {\makecell{\textsc{Qwen3-1.7B}\\\textsc{+ SFT + RL}\\(ours)}}{anchor=west, xshift=5pt}

\ourmodelpoint{rlzero17}{58.3}{24.8}
  {\makecell{\textsc{Qwen3-1.7B}\\\textsc{+ RL}}}{anchor=east, xshift=-5pt}
  
\draw[blue_arrow, shorten <= 1.5mm, shorten >= 1.5mm] (qwen17) -- (sft17);

\draw[green_arrow, shorten <= 1.5mm, shorten >= 1.5mm] (sft17) -- (ours17);

\draw[green_arrow, shorten <= 1.5mm, shorten >= 1.5mm] (qwen17) -- (rlzero17);

\addlegendimage{
  legend image code/.code={
    \draw[blue_arrow] (0cm,0cm) -- (0.38cm,0cm);
  }
}
\addlegendentry{SFT}

\addlegendimage{
  legend image code/.code={
    \draw[green_arrow] (0cm,0cm) -- (0.38cm,0cm);
  }
}
\addlegendentry{RL}

\end{axis}
\end{tikzpicture}
\caption{
Short-/long-context performance frontier.
}
\label{fig:performance_frontier}
\end{figure}

\paragraph{Our recipe shifts the short-/long-context performance frontier.}
Figure~\ref{fig:performance_frontier} shows that our recipe improves the trade-off relative to the base model while remaining competitive with stronger external long-context models without sacrificing short-context performance on average.
The SFT-only checkpoint already improves long-context ability while preserving short-context competence, and teacher-anchored RL adds further gains.
By contrast, applying the same RL objective directly to the base model (RL-Zero), without the SFT warm start, yields a worse trade-off, underscoring the role of Stage~1 in stabilizing long-horizon on-policy optimization.
The final model is competitive with larger 3B--4B models, including a YaRN-based context-extension model from the same family.
These external models serve as contextual reference points for the achieved performance frontier rather than controlled methodological comparisons, given differences in training data, compute, and adaptation procedures.
Appendix~\ref{sec:appendix-transfer-scaling} provides complementary generalization checks along two axes: family transfer to Gemma4-E2B and same-family scaling to Qwen3-4B.

\subsection{Dissecting the Training Recipe}
\label{sec:experiments:ablations}

We use matched ablations to isolate the contributions of the main components of our recipe.
We first examine the RL objective, separating outcome-reward optimization, reference-policy anchoring, and dense teacher guidance, and then study teacher choice and anchoring strength, cold-start initialization, and the short-/long-context data mixture.

\colorlet{refKLColor}{violet!55}

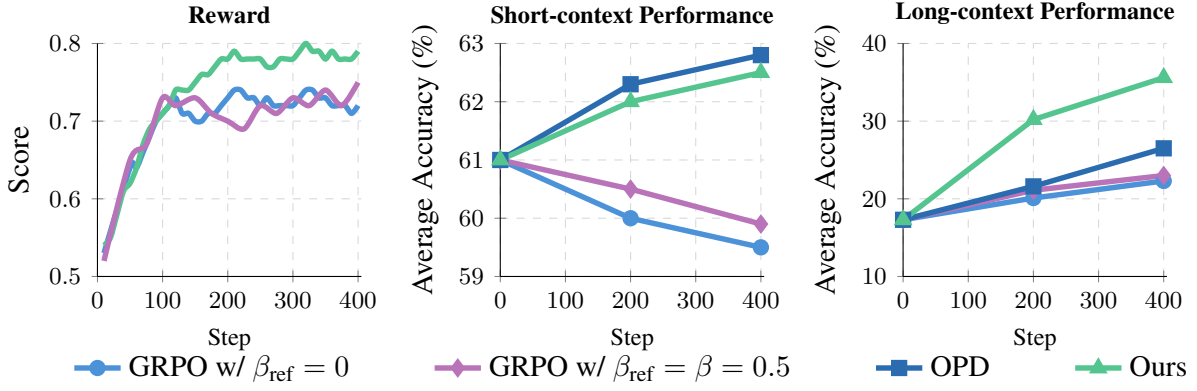
\begin{figure*}[!t]
    \centering
    \resizebox{\textwidth}{!}{
    \begin{tikzpicture}
    \begin{groupplot}[
        group style={
            group size=3 by 1,
            horizontal sep=1.8cm,
        },
        width=5cm,
        height=4.6cm,
        grid=major,
        grid style={dashed, gray!30},
        axis lines=left,
        axis line style={-}, 
        clip=false, 
        xlabel={Step},
        xmin=0, xmax=405,
        tick label style={font=\small},
        label style={font=\small},
        title style={font=\bfseries\small, at={(0.5, 0.97)}, anchor=south}, 
        every axis plot/.append style={line width=2pt},
        ylabel near ticks
    ]

    \nextgroupplot[
        title={Reward},
        ylabel={Score},
        ymin=0.5, ymax=0.8,
    ]
        \addplot[smooth, color=baselineColor] coordinates {
        (10, 0.53) (50, 0.64) (60, 0.64) (70, 0.66) (80, 0.68)
        (90, 0.70) (100, 0.71) (110, 0.72) (120, 0.73) (130, 0.71)
        (140, 0.71) (150, 0.70) (160, 0.70) (170, 0.71) (180, 0.71) 
        (190, 0.72) (200, 0.73) (210, 0.74) (220, 0.74) (230, 0.73) 
        (240, 0.73) (250, 0.72) (260, 0.73) (270, 0.72) (280, 0.72) 
        (290, 0.72) (300, 0.72) (310, 0.73) (320, 0.74) (330, 0.74) 
        (340, 0.73) (350, 0.73) (360, 0.72) (370, 0.72) (380, 0.72)
        (390, 0.71) (400, 0.72) };

        \addplot[smooth, color=ultraColor] coordinates {
            (10, 0.54) (20, 0.55) (30, 0.58) (40, 0.61) (50, 0.62)
            (60, 0.64) (70, 0.67) (80, 0.69) (90, 0.70) (100, 0.71)
            (110, 0.72) (120, 0.74) (130, 0.74) (140, 0.74) (150, 0.75)
            (160, 0.76) (170, 0.76) (180, 0.77) (190, 0.78) (200, 0.78)
            (210, 0.79) (220, 0.78) (230, 0.78) (240, 0.78) (250, 0.78)
            (260, 0.77) (270, 0.77) (280, 0.78) (290, 0.78) (300, 0.78)
            (310, 0.79) (320, 0.80) (330, 0.79) (340, 0.79) (350, 0.78)
            (360, 0.79) (370, 0.78) (380, 0.78) (390, 0.78) (400, 0.79) };

        \addplot[smooth, color=refKLColor] coordinates {
         (10, 0.52) (50, 0.65) (75, 0.67) (100, 0.73) (120, 0.72) (150, 0.73) (175, 0.71) (200, 0.7) (225, 0.69) (250, 0.72) (275, 0.71) (300, 0.73) (325, 0.72) (350, 0.74) (375, 0.72) (400, 0.75)
         };
        
    \nextgroupplot[
        title={Short-context Performance},
        ylabel={Average Accuracy (\%)},
        ymin=59, ymax=63,
        legend style={at={(0.5,-0.35)}, anchor=north, legend columns=-1, draw=none, fill=none, /tikz/every even column/.append style={column sep=1cm}},
    ]
        \addplot[color=baselineColor, mark=*] coordinates {(0, 61.0) (200, 60.0) (400, 59.5)};
        \addlegendentry{GRPO w/ $\beta_{\text{ref}} = 0$}

        \addplot[color=refKLColor, mark=diamond*] coordinates {(0, 61.0) (200, 60.5) (400, 59.9)};
        \addlegendentry{GRPO w/ $\beta_{\text{ref}} = \beta = 0.5$}

        \addplot[color=baselineColorv2, mark=square*] coordinates {(0, 61.0) (200, 62.3) (400, 62.8)};
        \addlegendentry{OPD}

        \addplot[color=ultraColor, mark=triangle*] coordinates {(0, 61.0) (200, 62.0) (400, 62.5)};
        \addlegendentry{Ours}
        
    \nextgroupplot[
        title={Long-context Performance},
        ylabel={Average Accuracy (\%)},
        ymin=10, ymax=40
    ]
        \addplot[color=baselineColor, mark=*] coordinates {(0, 17.3) (200, 20.1) (400, 22.3)};
        
        \addplot[color=refKLColor, mark=diamond*] coordinates {(0, 17.3) (200, 21.1) (400, 23.0)};

        \addplot[color=baselineColorv2, mark=square*] coordinates {(0, 17.3) (200, 21.6)(400, 26.5)};

        \addplot[color=ultraColor, mark=triangle*] coordinates {(0, 17.3) (200, 30.2) (400, 35.6)};
        
    \end{groupplot}
    \end{tikzpicture}}
\caption{Comparison of GRPO with and without reference KL~\citep{shao2024deepseekmathpushinglimitsmathematical}, OPD~\citep{agarwal2024onpolicy}, and our teacher-anchored RL objective. OPD has no reward trajectory because it does not optimize rewards.}
\label{fig:grpo_vs_opd_vs_ours}
\end{figure*}
\begin{figure*}[!t]
    \centering
    \resizebox{\textwidth}{!}{
    \begin{tikzpicture}
    \begin{groupplot}[
        group style={
            group size=3 by 1,
            horizontal sep=1.8cm,
        },
        width=5cm,
        height=4.6cm,
        grid=major,
        grid style={dashed, gray!30},
        axis lines=left,
        axis line style={-}, 
        clip=false, 
        xlabel={Step},
        xmin=0, xmax=405,
        tick label style={font=\small},
        label style={font=\small},
        title style={font=\bfseries\small, at={(0.5, 0.97)}, anchor=south}, 
        every axis plot/.append style={line width=2pt},
        ylabel near ticks
    ]

    \nextgroupplot[
        title={Reward},
        ylabel={Score},
        ymin=0.5, ymax=0.8,
    ]
        \addplot[smooth, color=ultraColor] coordinates {
            (10, 0.54) (20, 0.55) (30, 0.58) (40, 0.61) (50, 0.62)
            (60, 0.64) (70, 0.67) (80, 0.69) (90, 0.70) (100, 0.71)
            (110, 0.72) (120, 0.74) (130, 0.74) (140, 0.74) (150, 0.75)
            (160, 0.76) (170, 0.76) (180, 0.77) (190, 0.78) (200, 0.78)
            (210, 0.79) (220, 0.78) (230, 0.78) (240, 0.78) (250, 0.78)
            (260, 0.77) (270, 0.77) (280, 0.78) (290, 0.78) (300, 0.78)
            (310, 0.79) (320, 0.80) (330, 0.79) (340, 0.79) (350, 0.78)
            (360, 0.79) (370, 0.78) (380, 0.78) (390, 0.78) (400, 0.79) };
            
        \addplot[smooth, color=baselineColor] coordinates {
            (10, 0.51) (20, 0.52) (30, 0.53) (40, 0.55) (50, 0.57)
            (60, 0.60) (70, 0.61) (80, 0.63) (90, 0.64) (100, 0.64)
            (110, 0.65) (120, 0.67) (130, 0.67) (140, 0.68) (150, 0.68)
            (160, 0.69) (170, 0.70) (180, 0.67) (190, 0.68) (200, 0.69)
            (210, 0.70) (220, 0.71) (230, 0.70) (240, 0.71) (250, 0.71)
            (260, 0.72) (270, 0.71) (280, 0.7) (290, 0.69) (300, 0.7)
            (310, 0.7) (320, 0.71) (330, 0.69) (340, 0.68) (350, 0.69)
            (360, 0.68) (370, 0.69) (380, 0.70) (390, 0.69) (400, 0.70) };

    \nextgroupplot[
        title={Short-context Performance},
        ylabel={Average Accuracy (\%)},
        ymin=61, ymax=62.5,
        legend style={at={(0.5,-0.35)}, anchor=north, legend columns=-1, draw=none, fill=none, /tikz/every even column/.append style={column sep=1cm}},
    ]
        \addplot[color=ultraColor, mark=triangle*] coordinates {(0, 61.0) (200, 62.0) (400, 62.5)};
        \addlegendentry{External teacher}
        \addplot[color=baselineColor, mark=*] coordinates {(0, 61.0) (200, 61.5) (400, 61.7)};
        \addlegendentry{Self-teacher}

    \nextgroupplot[
        title={Long-context Performance},
        ylabel={Average Accuracy (\%)},
        ymin=10, ymax=40
    ]
        \addplot[color=ultraColor, mark=triangle*] coordinates {(0, 17.3) (200, 30.2) (400, 35.6)};
        \addplot[color=baselineColor, mark=*] coordinates {(0, 17.3) (200, 20.7) (400, 25.0)};
        
    \end{groupplot}
    \end{tikzpicture}}
    \caption{Comparison of external-teacher and self-teacher variants of our teacher-anchored RL objective, showing reward trajectories and short-/long-context evaluation performance.}
    \label{fig:teacher_vs_self_distillation}
\end{figure*}
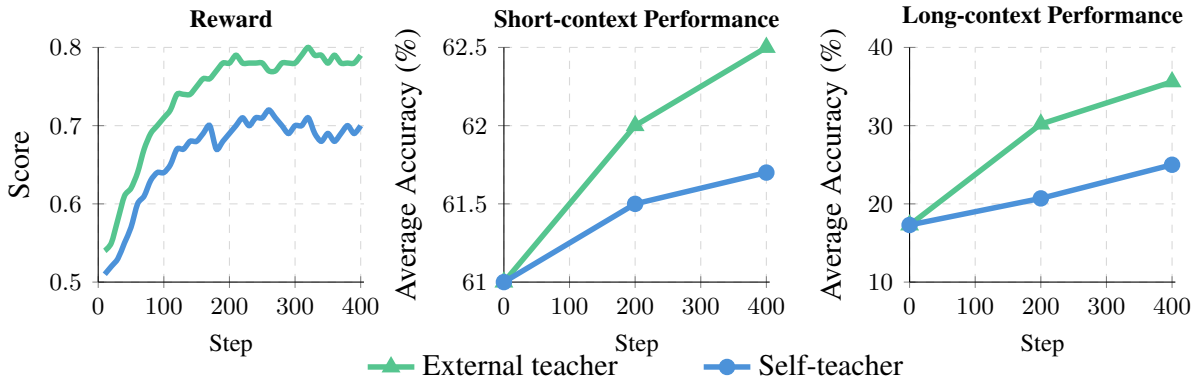
\begin{figure}[!t]
\centering
\begin{tikzpicture}
\begin{axis}[
  width=0.38\textwidth,
  height=0.23\textwidth,
  scale only axis,
  xmin=20, xmax=37,
  ymin=58.5, ymax=64,
  xtick={20,25,30,35,40,45},
  ytick={59,60,61,62,63,64},
  grid=major,
  major grid style={draw=black!10, line width=0.3pt, dashed},
  axis x line*=bottom,
  axis y line*=left,
  axis line style={draw=black!65, line width=0.7pt},
  tick style={draw=black!55, line width=0.6pt},
  tick align=outside,
  ticklabel style={font=\scriptsize, text=black!85},
  label style={font=\small, text=black!90},
  xlabel={Long-context Performance},
  ylabel={Short-context Performance},
  clip=false,
]

\addplot[
  color=baselineColorv2,
  line width=1.4pt,
  mark=*,
  mark size=2.4pt,
] coordinates {
  (22.3, 59.5)
  (25.3, 60.8)
  (28.6, 61.5)
  (32.6, 62.1)
  (35.6, 62.5)
};

\addplot[
  only marks,
  mark=*,
  mark size=3.0pt,
  color=ultraColor,
] coordinates {(35.6, 62.5)};

\node[font=\small, anchor=center, yshift=-9.5pt, text=black!90] at (axis cs:22.3,59.5) {$\beta=0$};
\node[font=\small, anchor=center, yshift=10pt, text=black!90] at (axis cs:25.3,60.8) {$\beta=0.1$};
\node[font=\small, anchor=center, yshift=10pt, text=black!90] at (axis cs:28.6,61.5) {$\beta=0.25$};
\node[font=\small, anchor=center, yshift=10pt, text=black!90] at (axis cs:32.6,62.1) {$\beta=0.4$};
\node[font=\small, anchor=center, xshift=-5pt, yshift=15pt, text=ultraColor] at (axis cs:35.6,62.5) {$\beta=0.5/0.7$};
\end{axis}
\end{tikzpicture}
\caption{Effect of $\beta$ during the RL stage on the short-/long-context performance trade-off.}
\label{fig:beta_effect}
\end{figure}
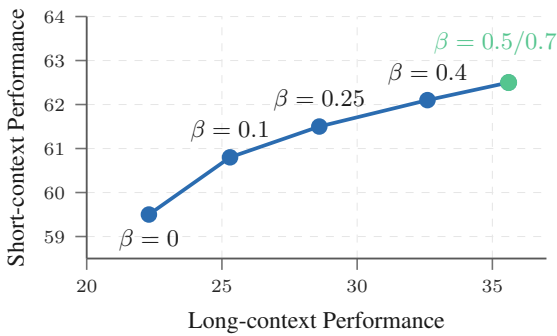
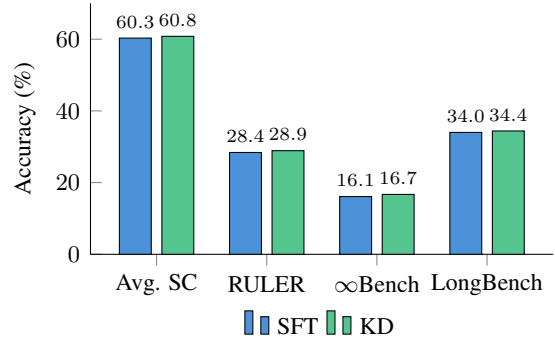
\begin{figure}[t]
    \centering
    \begin{tikzpicture}[trim axis left, trim axis right]
        \begin{axis}[
            axis lines*=left, 
            ybar=4pt,
            bar width=12pt,
            width=\linewidth,
            height=5.3cm,
            ylabel={Accuracy (\%)},
            label style={font=\small},
            symbolic x coords={Avg. SC, RULER, \infiniteBenchNS, LongBench},
            xtick=data,
            xticklabel style={rotate=0, anchor=center, font=\small, yshift=-7pt},
            yticklabel style={font=\small},
            legend style={
                at={(0.5,-0.2)},
                anchor=north,
                legend columns=2,
                font=\small,
                column sep=1pt,
                draw=none
            },
            ymin=0, ymax=70,
            enlarge x limits=0.2,
            nodes near coords,
            every node near coord/.append style={
                text=black,
                font=\scriptsize, 
                rotate=0, 
                /pgf/number format/fixed,
                /pgf/number format/fixed zerofill,
                /pgf/number format/precision=1
            },
        ]

        \addplot+[fill=baselineColor, draw=black] coordinates {
            (Avg. SC, 60.3) (\infiniteBenchNS, 16.1) (RULER, 28.4) (LongBench, 34.0)
        };

        \addplot+[fill=ultraColor, draw=black] coordinates {
            (Avg. SC, 60.8) (\infiniteBenchNS, 16.7) (RULER, 28.9) (LongBench, 34.4)
        };
        
        \legend{SFT, KD}
        \end{axis}
    \end{tikzpicture}
    \caption{Comparison of SFT and KD on short-context (SC) vs. long-context benchmarks.}
    \label{fig:sft_vs_kd}
\end{figure}
\begin{figure*}[!htbp]
    \centering
    \begin{tikzpicture}
        \begin{groupplot}[
            group style = {group size = 2 by 1, horizontal sep = 60pt},
            width = 7cm, 
            height = 5cm,
            axis x line*=bottom,
            axis y line*=left,
            xtick pos=bottom,
            ytick pos=left,
            grid=both,
            grid style=dashed,
            xlabel={Percent (\%) of long-context training data},
            ylabel={Accuracy (\%)},
            label style={font=\small},
            tick label style={font=\small},
            xmin=-0.01, xmax=1.01,
            xtick={0.0, 0.1, 0.25, 0.5, 0.75, 0.9, 1.0},
            xticklabels={0, 10, 25, 50, 75, 90, 100},
        ]
        
        \nextgroupplot[
            title = {\small\textbf{\textsc{Short-context Performance}}},
            title style={yshift=-1ex},
            ymin=54, ymax=70,
            ytick={50, 55, 60, 65, 70},
            legend style={
                at={(1.1,-0.3)},
                anchor=north,
                legend columns=2,
                draw=none,
            },
            legend cell align=left,
        ]

        \addplot[
            color=baselineColor,
            dashed,
            line width=1.5pt,
        ]{61.0};
        \addlegendentry{Qwen3-1.7B}

        \addplot[
            color=ultraColor,
            line width=1.5pt,
            mark=square*,
        ] coordinates {
            (1.0, 55.0)
            (0.9, 62.5)
            (0.75, 64.1)
            (0.5, 65.6)
            (0.25, 66.55)
            (0.1, 68.11)
            (0.0, 69.00)
        };
        \addlegendentry{Ours}

        \nextgroupplot[
            title = {\small\textbf{\textsc{Long-context Performance}}},
            title style={yshift=-1ex},
            ymin=13, ymax=40,
            ytick={15, 20, 25, 30, 35, 40},
        ]

        \addplot[
            color=baselineColor,
            dashed,
            line width=1.5pt,
        ]{17.3};

        \addplot[
            color=ultraColor,
            line width=1.5pt,
            mark=square*,
        ] coordinates {
            (1.0, 36.3)
            (0.9, 35.6)
            (0.75, 28.00)
            (0.5, 26.70)
            (0.25, 24.70)
            (0.1, 22.82)
            (0.0, 14.00)
        };

        \end{groupplot}   
    \end{tikzpicture}
    \caption{Accuracy on short- and long-context tasks across different short/long data mixes.}
    \label{fig:sc_lc_data_mix}
\end{figure*}
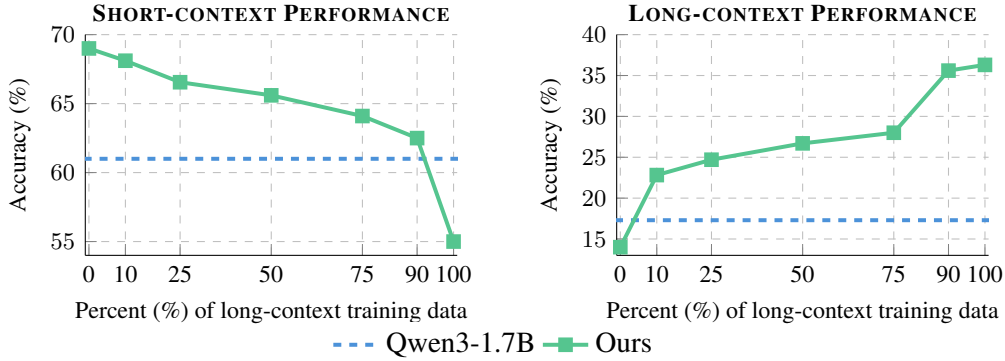

\paragraph{Teacher-anchored RL stabilizes on-policy optimization and improves long-context reasoning.}
Figure~\ref{fig:grpo_vs_opd_vs_ours} compares GRPO, GRPO with reference-policy KL regularization ($\beta_{\mathrm{ref}}=0.5$), OPD, and our teacher-anchored RL objective under the same SFT warm start, one-epoch RL budget, and rollout settings from Section~\ref{sec:longcontext}.
This matched-budget comparison isolates sparse outcome optimization, reference-policy anchoring, dense teacher guidance, and their combination.
GRPO improves reward but exhibits a noisier reward trajectory, underscoring the difficulty of optimizing long-context behavior from sparse trajectory-level rewards.
Adding reference-policy KL yields only modest improvements over GRPO, highlighting that reference-policy KL primarily constrains policy drift, whereas teacher KL provides dense supervision toward a stronger policy.
OPD provides dense token-level supervision and smoother updates, but yields smaller long-context gains because it does not optimize task rewards.
Our teacher-anchored objective combines outcome-reward optimization with dense teacher guidance, achieving the highest final reward among reward-optimizing methods, the strongest long-context performance among the matched variants, and essentially unchanged short-context accuracy.

\paragraph{Teacher quality and compute trade-off in teacher-anchored RL.}
Figure~\ref{fig:teacher_vs_self_distillation} compares our teacher-anchored RL stage using either a stronger external teacher or self-distillation using the current model as a privileged-context teacher conditioned on the reference answer \citep{zhao2026selfdistilledreasoneronpolicyselfdistillation}.
The teacher introduces additional post-training compute through forward passes on student-generated trajectories, but does not generate rollouts.
Self-distillation provides a lower-cost alternative that avoids a separate teacher model and improves long-context performance to 25.0, compared with 35.6 using the 32B teacher.
Thus, teacher quality can be traded against available post-training compute: self-teacher anchoring retains part of the benefit, while a stronger teacher yields substantially larger gains.
Importantly, this overhead is limited to post-training; deployment uses only the resulting student model and incurs no additional inference cost.

\paragraph{Tuning $\beta$: Regularization vs.\ Exploration.}
Figure~\ref{fig:beta_effect} shows the effect of the teacher-anchoring strength $\beta$ during RL.
At $\beta=0$, the objective reduces to GRPO and performs worst, highlighting the difficulty of sparse-reward optimization without teacher guidance.
Larger $\beta$ improves both short- and long-context performance by regularizing model-generated trajectories.
Gains peak at $\beta=0.5$ and remain comparable at $\beta=0.7$, suggesting that moderate anchoring is sufficient and that further increases yield diminishing returns.
Overall, $\beta$ controls the trade-off between reward-driven exploration and teacher-anchored stability, consistent with \citet{agarwal2024onpolicy}.

\paragraph{Cold-start initialization: SFT vs.\ KD.}
Figure~\ref{fig:sft_vs_kd} compares SFT and KD as off-policy cold-start initialization strategies. The two approaches yield similar performance on both short- and long-context benchmarks, with KD providing only marginal improvements while requiring teacher inference when logits are not cached. We therefore adopt SFT as the default initialization. We additionally evaluate continued SFT from the resulting checkpoint as a matched off-policy-only baseline, using the same 10/90 data mixture and context length with a post-warm-up training budget comparable to that of the subsequent RL stage. Continued SFT leaves downstream performance essentially unchanged relative to the initial SFT checkpoint (SC Avg.: 60.3$\rightarrow$60.2; RULER: 28.4$\rightarrow$28.7; $\infty$Bench: 16.1$\rightarrow$16.2; LongBench: 34.0$\rightarrow$33.8), showing that the final gains cannot be explained by additional supervised exposure alone.

\paragraph{Balancing short- and long-context data.}
We ablate the short-/long-context data mix under a fixed token budget. Figure~\ref{fig:sc_lc_data_mix} shows that short-context accuracy is robust to the mix: only 10\% short-context data is enough to preserve performance. Long-context accuracy, however, drops as short-context data increases. We therefore use a 10/90 short-/long-context mix, which preserves short-context capabilities on average while maximizing long-context gains.

\section{Related Work}

\paragraph{Off-Policy vs.\ On-Policy Methods.}
Alignment methods for language models include off-policy methods, such as supervised fine-tuning, distillation, DPO~\citep{rafailov2024directpreferenceoptimizationlanguage}, and KTO~\citep{ethayarajh2024ktomodelalignmentprospect}, which optimize fixed demonstrations or preferences.
These methods are stable and compute-efficient, but can suffer from train--inference mismatch and exposure bias~\citep{bengio2015scheduledsampling,ranzato2016sequenceleveltrainingrecurrent}.
By contrast, on-policy methods optimize samples from the current policy, including PPO~\citep{ppo}, GRPO~\citep{shao2024deepseekmathpushinglimitsmathematical}, DAPO~\citep{yu2025dapo}, Dr.~GRPO~\citep{liu2025understanding}, and on-policy distillation methods~\citep{agarwal2024onpolicy,yang2025qwen3technicalreport,zhao2026selfdistilledreasoneronpolicyselfdistillation,hübotter2026reinforcementlearningselfdistillation}.
This distinction is especially important for long-horizon generation, where local errors compound over extended rollouts.
Our recipe combines these strengths through an off-policy cold start followed by teacher-anchored on-policy optimization.

\paragraph{Long-Context Adaptation and Post-Training.}
Prior work on long-context LLMs has focused either on extending context windows through architectural or positional modifications~\citep{su2021roformer,chen2023extending,peng2024yarn}, or on adapting models with static long-context data~\citep{bai-etal-2024-longalign,chen2024longloraefficientfinetuninglongcontext}.
While effective, static supervision does not train models on their own long-horizon failure modes and is often centered on narrow formats such as retrieval or summarization, despite the need for robust distant-dependency reasoning~\citep{liu-etal-2024-lost,bai-etal-2024-longbench,hsieh2024ruler}.
Our approach addresses this by using broader supervision from \longblocks\ together with optimization on model-generated trajectories.

\paragraph{Dense Guidance for Long-Context Alignment.}
Recent work has begun to explore on-policy alignment for long-form generation and reasoning~\citep{bai2024longwriterunleashing10000word,zhang-etal-2025-longreward,wan2025qwenlongl1longcontextlargereasoning}, showing that long-context post-training can benefit from reward-driven or preference-based optimization.
Process-supervised reward models can densify sparse rewards~\citep{lightman2023lets,khalifa2025process}, but require reliable step-level supervision, which is scarce and difficult to define uniformly across retrieval, summarization, classification, and multilingual document QA.
In parallel, RL--distillation and dense-guidance methods use token-level supervision to stabilize or complement on-policy learning~\citep{agarwal2024onpolicy,yang2025qwen3technicalreport,hübotter2026reinforcementlearningselfdistillation,zhao2026selfdistilledreasoneronpolicyselfdistillation}.
However, their role in long-context alignment remains underexplored when sparse rewards, long rollouts, and short-context retention interact.
We study this regime by pairing GRPO-style reward optimization with an on-policy teacher anchor, yielding a recipe for stable long-context alignment.

\section{Conclusion}
We studied how to align LLMs for long-context reasoning by combining off-policy warm-up, sparse-reward on-policy RL, and on-policy distillation.
Our results show that these ingredients address complementary failure modes: SFT stabilizes long-horizon rollouts, GRPO adapts the model to its own generations, and teacher anchoring provides dense guidance for sparse-reward optimization.

We introduced a two-stage recipe, supported by \longblocks, that improves long-context performance while preserving short-context capabilities.
Ablations show that the gains come not from KL regularization alone, but from anchoring reward optimization to a stronger teacher.
Overall, outcome-level rewards and dense teacher guidance offer a practical path toward stable long-context reasoning.

\section*{Limitations} 
Due to computational constraints, we prioritized component-wise informative ablation studies.
Appendix~\ref{sec:appendix-transfer-scaling} provides initial evidence across one additional model family and one larger same-family student, but broader architecture and scale sweeps remain future work.
Our ablations compare component-matched variants rather than exhaustively tuned alternatives such as preference optimization.
We focus on teacher-anchored RL, whose effectiveness depends on teacher and reward quality.
However, at very long context lengths, on-policy training is costly due to multiple rollouts and teacher forward passes, motivating cached logits, selective distillation, or lower-cost teacher anchoring.
Finally, \longblocks is synthetic and LLM-filtered and may inherit artifacts or biases from its source corpora, generator, and verifier.
Generation and filtering may systematically favor examples, reasoning patterns, or answer styles aligned with the models used in the pipeline, while multilingual coverage does not imply uniform quality across languages, especially for lower-resource languages.
Although task-family balancing, deduplication, benchmark-overlap filtering, and judge validation mitigate these risks, they cannot eliminate them entirely.
We therefore leave fine-grained analysis of these effects to future work.

\section*{Ethical Considerations}
All source artifacts are used for research purposes in accordance with their licenses, terms of use, and access conditions.
\longblocks\ is released under CC BY-SA 4.0 where redistribution is permitted.
For non-redistributable research content, we provide reconstruction instructions in the dataset card rather than redistributing the underlying documents. 
The dataset card also documents these limitations and provides recommended uses for the dataset.
Although we use public or pre-filtered sources and apply cleaning, deduplication, and LLM-based verification, \longblocks\ may still inherit artifacts, biases, offensive content, or personally identifying information from its sources, generator, or verifier.
\longblocks\ is intended for research on long-context post-training.

\section*{Acknowledgments}
We thank the members of the SARDINE lab for their useful and constructive comments.
This work was supported by the project DECOLLAGE (ERC-2022-CoG 101088763), by the Portuguese Recovery and Resilience Plan through project C645008882-00000055 (Center for Responsible AI), and by FCT/MECI through national funds and when applicable co-funded EU funds under UID/50008: Instituto de Telecomunicações.

\bibliography{custom}

\appendix
\newpage
\section{Multilingual Synthetic Data Generation for Long Context}
\label{appendix:synthetic-qa}

\subsection{Generation Prompts}
We present the prompts used with \texttt{Qwen3-Next-80B-A3B-Thinking} to generate multilingual QA pairs. We discard incomplete, unsupported, weakly grounded, or locally answerable examples.

\begin{figure}[!ht]
\centering
\begin{AIbox}{Summary Question Prompt}

You are given the following document in \verb|{language}|:

\begin{verbatim}
{document}
\end{verbatim}

Your task is to create one \textbf{high-quality summary question} about the document and its corresponding answer.

\vspace{1em}
\textbf{Instructions:}
\begin{itemize}
    \item The question must require \textbf{synthesizing information from the entire document}.
    \item The answer must provide a \textbf{concise but complete summary} that captures all major themes, ideas, and relationships, reflecting a full and accurate understanding of the text.
    \item Your output must follow exactly the format below and should be in \verb|{language}|.
    \item Do not include anything else.
\end{itemize}

\textbf{Output Format:}
\begin{verbatim}
Question: <question>
Answer: <answer>
\end{verbatim}
\end{AIbox}
\end{figure}

\begin{figure}[!ht]
\centering
\begin{AIbox}{Information-Retrieval Question Prompt}

You are given the following document in \verb|{language}|:

\begin{verbatim}
{document}
\end{verbatim}

Your task is to create one \textbf{high-quality information-retrieval question} about the document and its corresponding answer.

\vspace{1em}
\textbf{Instructions:}
\begin{itemize}
    \item The question must require locating and \textbf{extracting information from multiple parts} of the document.
    \item Ensure that it can only be answered by someone who has \textbf{thoroughly read and understood the full text}.
    \item The answer must provide the \textbf{precise information requested}, directly grounded in the document.
    \item Your output must follow exactly the format below and should be in \verb|{language}|.
    \item Do not include anything else.
\end{itemize}

\textbf{Output Format:}
\begin{verbatim}
Question: <question>
Answer: <answer>
\end{verbatim}
\end{AIbox}
\end{figure}

\begin{figure}[!ht]
\centering
\begin{AIbox}{Multi-Hop Reasoning Question Prompt}

You are given the following document in \verb|{language}|:

\begin{verbatim}
{document}
\end{verbatim}

Your task is to create one \textbf{high-quality multi-hop reasoning question} about the document and its corresponding answer.

\vspace{1em}
\textbf{Instructions:}
\begin{itemize}
    \item The question must require \textbf{reasoning across multiple parts} of the document.
    \item It should combine information from \textbf{different sections, themes, or ideas}, requiring logical synthesis of the full text.
    \item The answer must clearly \textbf{explain the reasoning steps} or connections involved.
    \item Your output must follow exactly the format below and should be in \verb|{language}|.
    \item Do not include anything else.
\end{itemize}

\textbf{Output Format:}
\begin{verbatim}
Question: <question>
Answer: <answer>
\end{verbatim}
\end{AIbox}
\end{figure}

\begin{figure}[!ht]
\centering
\begin{AIbox}{Comprehensive Question Prompt}

You are given the following document in \verb|{language}|:

\begin{verbatim}
{document}
\end{verbatim}

Your task is to create one \textbf{high-quality question} and its corresponding answer.

\vspace{1em}
\textbf{Instructions:}
\begin{itemize}
    \item The question must require information from the \textbf{entire document}, not just a small part.
    \item It should involve \textbf{multiple aspects or themes} so that it can only be answered by someone who has fully read and understood the document.
    \item Your output must follow exactly the format below and should be in \verb|{language}|.
    \item Do not include anything else.
\end{itemize}

\textbf{Output Format:}
\begin{verbatim}
Question: <question>
Answer: <answer>
\end{verbatim}
\end{AIbox}
\end{figure}

\begin{figure}[!ht]
\centering
\begin{AIbox}{Classification Question Prompt}

You are given the following document in \verb|{language}|:

\begin{verbatim}
{document}
\end{verbatim}

Your task is to create one \textbf{high-quality classification question} about the document and its corresponding answer.

\vspace{1em}
\textbf{Instructions:}
\begin{itemize}
    \item The question must require \textbf{identifying or inferring categories, labels, or sentiment} from content spread across the entire document.
    \item Ensure that the classification \textbf{cannot be made without reading and understanding the full text}.
    \item The answer must clearly provide the correct classification or sentiment label and show that it is grounded in the document's overall content.
    \item Your output must be in \verb|{language}| and follow the format below exactly.
    \item Do not include anything else.
\end{itemize}

\textbf{Output Format:}
\begin{verbatim}
Question: <question>
Answer: <answer>
\end{verbatim}
\end{AIbox}
\end{figure}

\begin{table*}[!ht]
\centering
\small
\renewcommand{\arraystretch}{1.15}
\begin{tabular*}{\textwidth}{@{\extracolsep{\fill}}lcccll@{}}
\toprule
\textbf{Dataset} &
\textbf{Scale} &
\textbf{Max Context} &
\textbf{Multilingual} &
\textbf{Coverage} &
\textbf{Supervision Focus} \\
\midrule
LongAlpaca &
12,000 &
100K &
{\color{crossred}$\times$} No &
Long-document QA &
Instruction following \\

LongAlign &
9,888 &
64K &
{\color{crossred}$\times$} No &
Long-document QA &
Instruction following \\

\longblocks &
193,219 &
128K &
{\color{tickgreen}$\checkmark$} Yes &
Multilingual documents and code &
Long-context reasoning \\
\bottomrule
\end{tabular*}
\caption{Comparison between \longblocks\ and representative long-context instruction datasets.}
\label{tab:longblocks-comparison}
\end{table*}

\FloatBarrier

\begin{figure*}[!t]
    \centering
    \hspace*{1.5em}
    \begin{tikzpicture}[trim axis left, trim axis right]
        \begin{axis}[
            axis lines*=left,
            ybar stacked,
            bar width=10pt,
            width=1\linewidth,
            height=7.5cm,
            enlarge x limits=0.02,
            ylabel={Samples (logarithmic scale)},
            ylabel style={font=\small, yshift=-5pt},
            yticklabel style={font=\small},
            symbolic x coords={en,de,fr,es,it,el,sv,ja,nl,uk,ru,pt,hu,zh,hi,pl,ar,ca,ro,ko,sk,tr,fi,mt,hr,sl,no,gl,da,ga,lt},
            xtick=data,
            xticklabels={en,de,fr,es,it,el,sv,ja,nl,uk,ru,pt,hu,zh,hi,pl,ar,ca,ro,ko,sk,tr,fi,mt,hr,sl,no,gl,da,ga,lt},
            xticklabel style={font=\scriptsize, align=center, text height=1.5ex, text depth=.25ex, anchor=north},
            ymode=log,
            log basis y={10},
            ymin=1, ymax=100000,
            scaled y ticks=false,
            ytick={1,10,100,1000,10000,100000},
            yticklabels={$10^0$, $10^1$, $10^2$, $10^3$, $10^4$, $10^5$},
            legend style={
                at={(0.5,-0.1)},
                anchor=north,
                legend columns=6,
                font=\scriptsize,
                column sep=2pt,
                draw=none,
                /tikz/every even column/.append style={column sep=2pt}
            },
            clip=true,
        ]

        \addplot+[fill=violet!42!white, draw=black] coordinates {
            (en,1) (de,924) (fr,719) (es,470) (it,235)
            (el,1052) (sv,41) (ja,1929) (nl,93) (uk,1094)
            (ru,924) (pt,615) (hu,329) (zh,161) (hi,350)
            (pl,340) (ar,344) (ca,209) (ro,170) (ko,142)
            (sk,73) (tr,59) (fi,55) (mt,42) (hr,40)
            (sl,37) (no,35) (gl,19) (da,15) (ga,6) (lt,5)
        };

        \addplot+[fill=baselineColor!92!white, draw=black] coordinates {
            (en,63243) (de,41097) (fr,3477) (es,1) (it,1)
            (el,1) (sv,1) (ja,1) (nl,1) (uk,1)
            (ru,1) (pt,1) (hu,1) (zh,1) (hi,1)
            (pl,1) (ar,1) (ca,1) (ro,1) (ko,1)
            (sk,1) (tr,0.000001) (fi,1) (mt,1) (hr,1)
            (sl,1) (no,1) (gl,1) (da,1) (ga,1) (lt,1)
        };

        \addplot+[fill=ultraColor!92!white, draw=black] coordinates {
            (en,1) (de,7575) (fr,17172) (es,6471) (it,3156)
            (el,1919) (sv,1840) (ja,1) (nl,774) (uk,1)
            (ru,1) (pt,1) (hu,5) (zh,171) (hi,1)
            (pl,1) (ar,1) (ca,1) (ro,1) (ko,1)
            (sk,1) (tr,0.000001) (fi,1) (mt,1) (hr,1)
            (sl,1) (no,1) (gl,1) (da,1) (ga,1) (lt,1)
        };

        \addplot+[fill=baselineColorv2!55!white, draw=black] coordinates {
            (en,13963) (de,1) (fr,1) (es,1) (it,1)
            (el,1) (sv,1) (ja,1) (nl,1) (uk,1)
            (ru,1) (pt,1) (hu,1) (zh,1) (hi,1)
            (pl,1) (ar,1) (ca,1) (ro,1) (ko,1)
            (sk,1) (tr,0.000001) (fi,1) (mt,1) (hr,1)
            (sl,1) (no,1) (gl,1) (da,1) (ga,1) (lt,1)
        };

        \addplot+[fill=orange!55!brown!65!white, draw=black] coordinates {
            (en,8520) (de,294) (fr,804) (es,1) (it,320)
            (el,1) (sv,109) (ja,1) (nl,389) (uk,1)
            (ru,1) (pt,210) (hu,164) (zh,169) (hi,1)
            (pl,8) (ar,1) (ca,1) (ro,1) (ko,1)
            (sk,1) (tr,0.000001) (fi,1) (mt,1) (hr,1)
            (sl,1) (no,1) (gl,1) (da,1) (ga,1) (lt,1)
        };

        \addplot+[fill=teal!55!baselineColor, draw=black] coordinates {
            (en,2824) (de,1) (fr,1) (es,1) (it,1)
            (el,1) (sv,53) (ja,1) (nl,63) (uk,1)
            (ru,1631) (pt,1) (hu,263) (zh,36) (hi,1)
            (pl,1) (ar,77) (ca,1) (ro,1) (ko,1)
            (sk,1) (tr,36) (fi,1) (mt,1) (hr,1)
            (sl,1) (no,1) (gl,1) (da,1) (ga,1) (lt,1)
        };

        \legend{
            Wikipedia,
            Institutional Books,
            FineWeb2-HQ,
            ArXiv,
            PGBooks,
            StackExchange
        }
        \end{axis}
    \end{tikzpicture}
    \caption{
    Language-wise composition of the corpora used to generate multilingual synthetic long-context data for \longblocks. Each stacked bar corresponds to one natural language, and colored segments denote the number of samples contributed by each source. Programming and markup language examples from Stack-Edu are approximately balanced across languages and omitted from the figure for readability.
    }
    \label{fig:dataset-language-coverage}
\end{figure*}
\begin{figure}[!ht]
\centering
\begin{AIbox}{LLM Judge Verification Prompt}

You will receive a \textbf{question}, a \textbf{response} generated by an LLM, and a \textbf{ground-truth answer}. 

Your task is to determine whether the response is \textbf{accurate} according to the ground-truth.

\vspace{1em}
\textbf{Instructions:}
\begin{itemize}
    \item Respond with \textbf{only a single integer}:
    \begin{itemize}
        \item \textbf{1} → The response is accurate and aligns with the ground-truth.
        \item \textbf{0} → The response is inaccurate or does not match the ground-truth.
    \end{itemize}
    \item Do not include any explanations, text, or commentary—just the integer.
\end{itemize}

\textbf{Input Format:}
\begin{verbatim}
Question: {question}
Response: {response}
Ground-truth: {ground-truth}
\end{verbatim}

\textbf{Output Format:}
\begin{verbatim}
Answer: <int>
\end{verbatim}
\end{AIbox}
\end{figure}

\begin{table*}[!t]
\centering
\small
\setlength{\tabcolsep}{12pt}
\begin{tabular}{lccccc}
\toprule
\textbf{Comparison} & \textbf{Examples} & \textbf{Agreement (\%)} & \textbf{$\kappa$} & \textbf{Pos. F1} & \textbf{FPR/FNR (\%)} \\
\midrule
Qwen3-4B vs. Human & 250 & 84.8 & 0.696 & 0.847 & 14.4 / 16.0 \\
Qwen3-4B vs. Qwen3-32B & 10{,}000 & 91.6 & 0.832 & 0.915 & 7.4 / 9.4 \\
Qwen3-4B vs. Gemma-3-27B & 10{,}000 & 87.0 & 0.740 & 0.866 & 10.0 / 15.9 \\
\bottomrule
\end{tabular}
\caption{Reliability analysis of the Qwen3-4B judge used for binary long-context rewards.}
\label{tab:reward-reliability}
\end{table*}

\subsection{Dataset Composition}
At generation time, we sample the same number of source documents for each of five task families: summary, multi-part information retrieval, multi-hop reasoning, comprehensive QA, and classification.
Table~\ref{tab:longblocks-comparison} compares \longblocks\ with two public long-context instruction datasets commonly used for long-context adaptation: LongAlpaca~\citep{chen2024longloraefficientfinetuninglongcontext} and LongAlign~\citep{bai-etal-2024-longalign}.
\longblocks\ provides reasoning-oriented long-context supervision at larger scale and with broader multilingual and source coverage.
Its generation templates encourage evidence aggregation, multi-hop reasoning, summarization, classification, and grounded long-form generation; for reasoning-focused examples, reference answers expose the relevant reasoning steps or cross-document connections.

\paragraph{Filtering and contamination controls.}
Starting from approximately 275,000 generated examples, prompt-side exact and fuzzy deduplication removes 21,978 examples, while quality filtering removes a further 57,301 malformed, weakly grounded, unsupported, or locally answerable examples.
Benchmark-overlap filtering removes approximately 2,500 additional examples, primarily from Wikipedia and ArXiv, yielding the final 193,219-example \longblocks\ corpus.
For fuzzy deduplication, we use NeMo Curator~\citep{nemo_curator} with its default MinHash--LSH configuration (24-character n-grams, 20 LSH bands, and 13 MinHashes per band).
To remove detectable benchmark overlap, we canonicalize \longblocks\ source documents and benchmark contexts by lowercasing, removing markup, and normalizing whitespace, then remove exact matches and fuzzy near-duplicates against RULER, OneRULER, LongBench, and \infiniteBench.

\paragraph{Language-wise composition of \longblocks.}
Figure~\ref{fig:dataset-language-coverage} summarizes the language composition of the generated dataset.
The dataset covers 31 natural languages, with contributions from heterogeneous sources including Wikipedia, Institutional Books, FineWeb2-HQ, ArXiv, PGBooks, and StackExchange.
High-resource languages receive contributions from multiple sources, while Wikipedia broadens coverage to lower-resource languages.

\subsection{Reward and Judge Reliability}
\paragraph{LLM-as-a-Judge Prompt for Reward Computation.}
During RL, \texttt{Qwen3-4B} is used as the LLM judge to compute binary rewards by comparing student responses against the ground-truth answers.

\paragraph{Reward reliability.}
We evaluate the LM judge used for long-context rewards with human validation and large-scale judge agreement.
We compare Qwen3-4B decisions with human annotations on a 250-example stratified set spanning the five \longblocks\ task families and balanced across Qwen3-4B reward labels.
We also compare Qwen3-4B with Qwen3-32B and Gemma-3-27B on balanced 10{,}000-example subsets with equal positive and negative decisions from the comparison judge, measuring agreement on both accepted and rejected cases.
We report agreement, Cohen's $\kappa$, positive-label F1, and false-positive/false-negative rates.
As shown in Table~\ref{tab:reward-reliability}, Qwen3-4B shows strong agreement with human annotations and with both alternative judges, including the different-family Gemma-3-27B judge, suggesting that the binary reward signal is consistent across human and model-based evaluations.

Human agreement is also consistent across the five task families ($n=50$ each): 88.0\% for summarization, 84.0\% for multi-part retrieval, 84.0\% for multi-hop reasoning, 86.0\% for comprehensive QA, and 82.0\% for classification. Because the human per-language subsets are too small for reliable estimates, we assess language-level robustness using the 10{,}000-example cross-judge comparisons: agreement with Qwen3-32B remains above 88\% across all reported language buckets, while agreement with Gemma-3-27B is at least 82\% in every bucket. We do not report a context-length breakdown because the validation examples are concentrated near the 128K limit, leaving insufficient variation for meaningful length-based stratification.

\section{Detailed Results}
\label{sec:appendix-detailed-results}

This appendix reports full benchmark score breakdowns, including run-to-run standard deviations, long-context results, and multilingual results.

\subsection{Run-to-Run Variability}
Evaluation with reasoning models can be computationally expensive and subject to nontrivial run-to-run variability under stochastic decoding. To account for this variability, we evaluate all main benchmark results over five runs using the shared decoding configuration described in Section~\ref{sec:experiments:setup}.
Main text reports the mean score across runs for readability, while this section reports the standard deviation across five runs for each benchmark. 
As shown in Table~\ref{tab:benchmark-std}, these results indicate that our post-training approach does not introduce additional variability, with standard deviations that remain broadly in line with those of reasoning models, including Qwen3-1.7B.

\begin{table}[!t]
\centering
\small
\setlength{\tabcolsep}{8pt}
\renewcommand{\arraystretch}{1.2}
\begin{tabular}{lcc}
\toprule
\textbf{Benchmark} & \textbf{Qwen3-1.7B} & \textbf{Ours} \\
\midrule
IFEval & 0.6608 & 0.7155 \\
IFBench & 0.9901 & 0.6487 \\
MMLU-Pro & 0.1441 & 0.1427 \\
GPQA $\blacklozenge$ & 1.9992 & 1.9499 \\
GSM8K & 0.4654 & 0.4461 \\
MATH-500 & 0.4775 & 0.3578 \\
HumanEval & 0.8627 & 0.7550 \\
HumanEval+ & 0.5561 & 0.4650 \\
\midrule
RULER & 0.3030 & 0.1500 \\
LongBench & 0.2550 & 0.2168 \\
$\infty$Bench & 0.3265 & 0.3918 \\
\bottomrule
\end{tabular}
\caption{Standard deviation across five evaluation runs for the main benchmark suite.}
\label{tab:benchmark-std}
\end{table}

\subsection{Benchmark Breakdowns}
Tables~\ref{tab:mmmlu}--\ref{tab:oneruler} provide the full multilingual and long-context benchmark breakdowns supporting the aggregate results in the main text.

\begin{table}[!ht]
\centering
\small
\setlength{\tabcolsep}{8pt}
\renewcommand{\arraystretch}{1.25}
\begin{tabular}{ccc}
\toprule
\textbf{Language} & \textbf{Qwen3-1.7B} & \textbf{Ours} \\
\midrule
\textit{de} & \textbf{80.8} & 79.2 \\
\textit{es} & 79.2 & \textbf{80.0} \\
\textit{fr} & \textbf{81.2} & 80.8 \\
\textit{ja} & 52.8 & \textbf{57.6} \\
\textit{ru} & \textbf{70.8} & \textbf{70.8} \\
\textit{zh} & 70.4 & \textbf{71.2} \\
\midrule
\textbf{Average} & 72.5 & \textbf{73.3} \\
\bottomrule
\end{tabular}
\caption{MGSM \citep{shi2022language} results by language.}
\label{tab:mgsm}
\end{table}

\begin{table}[!ht]
\centering
\small
\setlength{\tabcolsep}{8pt}
\renewcommand{\arraystretch}{1.25}
\begin{tabular}{ccc}
\toprule
\textbf{Language} & \textbf{Qwen3-1.7B} & \textbf{Ours} \\
\midrule
\textit{ar} & 53.2 & \textbf{54.1} \\
\textit{ca} & \textbf{60.6} & 60.5 \\
\textit{da} & \textbf{58.8} & 58.0 \\
\textit{de} & 61.1 & \textbf{62.8} \\
\textit{es} & 62.7 & \textbf{62.8} \\
\textit{fr} & \textbf{63.0} & 62.4 \\
\textit{hi} & 49.7 & \textbf{50.6} \\
\textit{hu} & \textbf{56.2} & \textbf{56.2} \\
\textit{it} & 62.7 & \textbf{62.9} \\
\textit{nl} & \textbf{61.3} & 61.0 \\
\textit{pt} & 62.7 & \textbf{63.5} \\
\textit{ro} & \textbf{61.2} & 60.5 \\
\textit{ru} & 61.4 & \textbf{61.6} \\
\textit{sv} & \textbf{60.5} & 60.3 \\
\textit{uk} & 58.5 & \textbf{60.2} \\
\textit{zh} & 55.4 & \textbf{55.8} \\
\midrule
\textbf{Average} & 59.3 & \textbf{59.6} \\
\bottomrule
\end{tabular}
\caption{MMMLU \citep{openai_mmmlu_2024} results by language.}
\label{tab:mmmlu}
\end{table}

\begin{table}[!ht]
\centering
\small
\setlength{\tabcolsep}{8pt}
\renewcommand{\arraystretch}{1.25}
\begin{tabular}{lcc}
\toprule
\textbf{Category} & \textbf{Qwen3-1.7B} & \textbf{Ours} \\
\midrule
\textit{Single-Document QA} & 34.3 & \textbf{39.2} \\
\textit{Multi-Document QA}  & 23.4 & \textbf{43.1} \\
\textit{Summarization}      & 20.0 & \textbf{20.2} \\
\textit{Few-shot Learning}  & 42.9 & \textbf{43.3} \\
\textit{Synthetic Tasks}    & \textbf{65.3} & 65.0 \\
\textit{Code Completion}    & 20.3 & \textbf{26.9} \\
\midrule
\textbf{Average}            & 34.4 & \textbf{39.6} \\
\bottomrule
\end{tabular}
\caption{LongBench \citep{bai-etal-2024-longbench}.}
\label{tab:longbench}
\end{table}

\begin{table}[!t]
\centering
\small
\setlength{\tabcolsep}{8pt}
\renewcommand{\arraystretch}{1.25}
\begin{tabular}{lcc}
\toprule
\textbf{Category} & \textbf{Qwen3-1.7B} & \textbf{Ours} \\
\midrule
\textit{R.PassKey}     & 13.9 & \textbf{100.0} \\
\textit{R.Number}      & 15.4 & \textbf{93.2} \\
\textit{Retrieve.KV}   & \textbf{0.0}  & \textbf{0.0} \\
\textit{En.Sum}        & 11.5 & \textbf{11.8} \\
\textit{En.QA}         & 0.1  & \textbf{8.1} \\
\textit{Zh.QA}         & 3.9  & \textbf{5.7} \\
\textit{En.MC}         & 18.8 & \textbf{30.1} \\
\textit{En.Dia}        & 0.5  & \textbf{1.2} \\
\textit{Code.Debug}    & 3.7  & \textbf{4.6} \\
\textit{Code.Run}      & 0.5  & \textbf{1.5} \\
\textit{Math.Calc}     & \textbf{6.0}  & \textbf{6.0} \\
\textit{Math.Find}     & 4.9  & \textbf{10.6} \\
\midrule
\textbf{Average}       & 6.6  & \textbf{22.7} \\
\bottomrule
\end{tabular}
\caption{$\infty$Bench \citep{zhang-etal-2024-bench}, including all benchmark categories. Many low scores (e.g., \textit{Retrieve.KV}, \textit{En.QA}, \textit{Code.Run}, \textit{Math.Calc}) reflect the difficulty of reasoning over extremely long contexts.}
\label{tab:inftybench}
\end{table}

\begin{table*}[!t]
\centering
\small
\setlength{\tabcolsep}{1.9pt}
\renewcommand{\arraystretch}{1.3}
\begin{tabular}{lccccccccccccccc}
\toprule
\small
\textbf{Model} & \textbf{Length} & \textbf{Avg.} & \textbf{S1} & \textbf{S2} & \textbf{S3} & \textbf{MK1} & \textbf{MK2} & \textbf{MK3} & \textbf{MV} & \textbf{MQ} & \textbf{VT} & \textbf{CWE} & \textbf{FWE} & \textbf{SQA} & \textbf{HQA} \\
\midrule

\multirow{6}{*}{\makecell{\textbf{Qwen3}\\\textbf{1.7B}}}
& 4K   & 87.4 & 100.0 & 100.0 & 99.8 & 100.0 & 95.4 & 98.6 & 99.1 & 100.0 & 99.0 & 91.8 & 75.7 & 42.0 & 34.2 \\
& 8K   & 84.2 & 100.0 & 100.0 & 99.6 & 100.0 & 92.4 & 92.6 & 97.1 & 99.8 & 97.3 & 73.9 & 76.3 & 32.2 & 34.0 \\
& 16K  & 81.0 & 100.0 & 100.0 & 100.0 & 99.4 & 81.8 & 93.8 & 93.2 & 99.6 & 92.5 & 46.4 & 88.7 & 29.5 & 28.6 \\
& 32K  & 72.0 & 100.0 & 100.0 & 99.8 & 89.6 & 74.6 & 45.2 & 89.0 & 96.5 & 88.8 & 19.4 & 79.9 & 28.0 & 25.6 \\
& 64K  & 51.5 & 100.0 & 97.2 & 90.6 & 55.0 & 3.2  & 0.2  & 64.2 & 84.3 & 55.2 & 7.5  & 61.6 & 27.8 & 22.3 \\
& 128K & 11.0 & 61.0  & 3.0  & 3.6  & 7.4  & 0.0  & 0.0  & 0.0  & 1.7  & 0.4  & 0.2  & 58.6 & 3.7  & 3.6  \\
\midrule

\multirow{6}{*}{\textbf{Ours}}
& 4K   & 88.3 & 100.0 & 100.0 & 99.8 & 99.8 & 97.2 & 99.6 & 91.7 & 100.0 & 99.1 & 90.5 & 82.7 & 47.0 & 40.2 \\
& 8K   & 85.4 & 100.0 & 100.0 & 99.8 & 99.6 & 93.2 & 97.4 & 94.7 & 99.9  & 99.1 & 83.5 & 73.1 & 34.4 & 36.0 \\
& 16K  & 84.7 & 100.0 & 100.0 & 97.0 & 99.6 & 90.0 & 97.4 & 92.3 & 100.0 & 91.5 & 77.5 & 88.2 & 32.5 & 35.0 \\
& 32K  & 80.4 & 100.0 & 100.0 & 98.2 & 99.6 & 91.6 & 80.4 & 91.3 & 98.7  & 80.7 & 71.2 & 67.1 & 32.1 & 33.8 \\
& 64K  & 65.6 & 100.0 & 100.0 & 98.6 & 85.6 & 32.0 & 22.4 & 92.7 & 95.0  & 70.4 & 48.3 & 44.6 & 35.7 & 27.2 \\
& 128K & 44.5 & 76.8  & 66.0  & 65.8 & 45.6 & 13.2 & 2.6  & 44.9 & 49.0  & 73.9 & 27.6 & 78.1 & 19.6 & 15.4 \\
\bottomrule
\end{tabular}
\caption{\textsc{RULER} \citep{hsieh2024ruler} breakdown across sequence lengths.}
\label{tab:ruler-breakdown}
\end{table*}

\begin{table*}[!t]
\centering
\small
\setlength{\tabcolsep}{4pt}
\renewcommand{\arraystretch}{1.25}
\begin{tabular}{c*{5}{c}*{5}{c}}
\toprule
\textbf{Language} &
\multicolumn{5}{c}{\textbf{Qwen3-1.7B}} &
\multicolumn{5}{c}{\textbf{Ours}} \\
\cmidrule(lr){2-6} \cmidrule(lr){7-11}
& \textbf{Avg.} & \textbf{8K} & \textbf{32K} & \textbf{64K} & \textbf{128K}
& \textbf{Avg.} & \textbf{8K} & \textbf{32K} & \textbf{64K} & \textbf{128K} \\
\midrule
\textit{de} & 63.3 & 99.0 & 84.8 & 69.5 & 0.1 & 75.1 & 99.0 & 97.5 & 74.1 & 29.6 \\
\textit{es} & 63.8 & 98.4 & 85.0 & 70.8 & 1.0 & 78.3 & 99.3 & 96.6 & 74.1 & 43.3 \\
\textit{fr} & 63.4 & 98.9 & 85.4 & 68.8 & 0.5 & 76.8 & 99.0 & 96.5 & 71.6 & 40.1 \\
\textit{it} & 62.8 & 98.9 & 84.5 & 67.0 & 0.8 & 75.3 & 99.0 & 97.1 & 71.1 & 34.1 \\
\textit{ja} & 64.9 & 99.0 & 86.4 & 74.2 & 0.0 & 73.9 & 99.0 & 96.3 & 79.4 & 21.0 \\
\textit{ko} & 83.8 & 98.9 & 86.4 & 74.9 & 7.9 & 86.5 & 98.8 & 96.4 & 77.9 & 72.9 \\
\textit{nl} & 62.4 & 98.1 & 83.4 & 68.1 & 0.1 & 74.9 & 98.8 & 97.6 & 71.5 & 31.5 \\
\textit{pt} & 63.9 & 98.4 & 85.1 & 71.6 & 0.5 & 77.8 & 98.9 & 97.1 & 76.0 & 39.1 \\
\textit{ru} & 63.7 & 99.0 & 85.4 & 68.9 & 1.5 & 78.3 & 98.9 & 96.3 & 74.0 & 43.8 \\
\textit{zh} & 64.7 & 98.5 & 86.4 & 74.0 & 0.0 & 74.1 & 98.6 & 94.8 & 78.8 & 24.0 \\
\midrule
\textbf{Average} & 65.7 & 98.7 & 85.3 & 70.8 & 1.2 & 77.1 & 98.9 & 96.6 & 74.9 & 37.9 \\
\bottomrule
\end{tabular}
\caption{OneRULER \citep{kim2025one} results by language.}
\label{tab:oneruler}
\end{table*}

\FloatBarrier

\subsection{Model Family Transfer and Scaling}
\label{sec:appendix-transfer-scaling}
We further assess the generality of our recipe along two axes. The \texttt{google/gemma-4-E2B-it} experiment tests transfer to a different small model family, while the Qwen3-4B experiment tests whether the same recipe remains effective for a larger student from the original family.
For the Gemma~4 E2B experiment, we use \texttt{google/gemma-4-31B-it} as a same-family teacher, preserving tokenizer compatibility for exact token-level KL. Thus, this experiment tests transfer of the training recipe across student families rather than cross-tokenizer teacher distillation.

The Gemma~4 E2B transfer setting follows the same trend as the primary Qwen3-1.7B experiment: long-context performance improves while short-context performance is retained. RULER 128K improves from 69.1 to 79.0, \infiniteBench improves from 15.6 to 30.7, and LongBench improves from 44.4 to 48.5.
The Qwen3-4B scaling setting shows the same pattern at a larger student scale. RULER 128K improves from 65.8 to 78.5, \infiniteBench improves from 13.0 to 26.6, and LongBench improves from 42.9 to 46.9, while short-context metrics are retained or improve on most benchmarks.

Together, the Gemma~4 E2B and Qwen3-4B results suggest that the recipe is not narrowly tied to the Qwen3-1.7B setup.

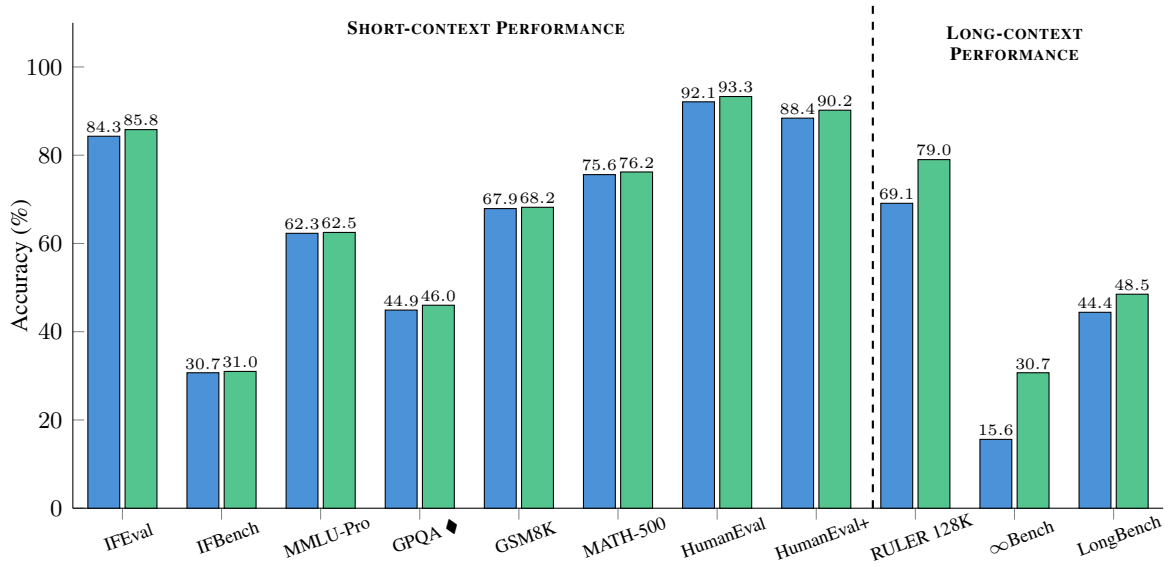
\begin{figure*}[!htbp]
    \centering
    \begin{tikzpicture}[trim axis left, trim axis right]
        \begin{axis}[
            axis lines*=left, 
            ybar=2pt,
            bar width=12pt,
            width=\linewidth,
            height=8cm,
            enlarge x limits=0.05,
            ylabel={Accuracy (\%)},
            ylabel style={font=\small, yshift=-10pt},
            yticklabel style={font=\small, xshift=0pt},
            symbolic x coords={
                IFEval, IFBench, MMLU-Pro, GPQA $\blacklozenge$, GSM8K, MATH-500, HumanEval, HumanEval+,
                RULER 128K, $\infty$Bench, LongBench
            },
            xtick=data,
            xticklabel style={rotate=20, anchor=center, yshift=-7.0pt, font=\scriptsize},
            legend style={
                at={(0.5,-0.3)},
                anchor=north,
                legend columns=2,
                font=\small,
                column sep=1pt,
                draw=none
            },
            ymin=0, ymax=110,
            clip=false,      
            nodes near coords,
            every node near coord/.append style={
                text=black,
                font=\tiny,
                /pgf/number format/fixed,
                /pgf/number format/fixed zerofill,
                /pgf/number format/precision=1,
                anchor=center,
                yshift=3.5pt
            },
        ]

        \addplot+[fill=baselineColor, draw=black] coordinates {
            (IFEval, 84.3) (IFBench, 30.7) (MMLU-Pro, 62.3) (GPQA $\blacklozenge$, 44.9) (GSM8K, 67.9) (MATH-500, 75.6) (HumanEval, 92.1) (HumanEval+, 88.4)
            (RULER 128K, 69.1) ($\infty$Bench, 15.6) (LongBench, 44.4)
        };

         \addplot+[fill=ultraColor, draw=black] coordinates {
            (IFEval, 85.8) (IFBench, 31.0) (MMLU-Pro, 62.5) (GPQA $\blacklozenge$, 46.0) (GSM8K, 68.2) (MATH-500, 76.2) (HumanEval, 93.3) (HumanEval+, 90.2)
            (RULER 128K, 79.0) ($\infty$Bench, 30.7) (LongBench, 48.5)
        };

        \draw[dashed, thick] ([xshift=-16pt]axis cs:RULER 128K,0) -- ([xshift=-16pt]axis cs:RULER 128K,115);

        \node[anchor=south, font=\scriptsize\bfseries] at ([xshift=25pt]axis cs:GPQA $\blacklozenge$, 105) {\textsc{Short-context Performance}};
        \node[anchor=center, font=\scriptsize\bfseries] at (axis cs:$\infty$Bench, 105) {\makecell{\textsc{Long-context}\\\textsc{Performance}}};

        \end{axis}
    \end{tikzpicture}
    \caption{Gemma4-E2B family-transfer benchmark performance.}
    \label{fig:gemma4-2b-main-results}
\end{figure*}
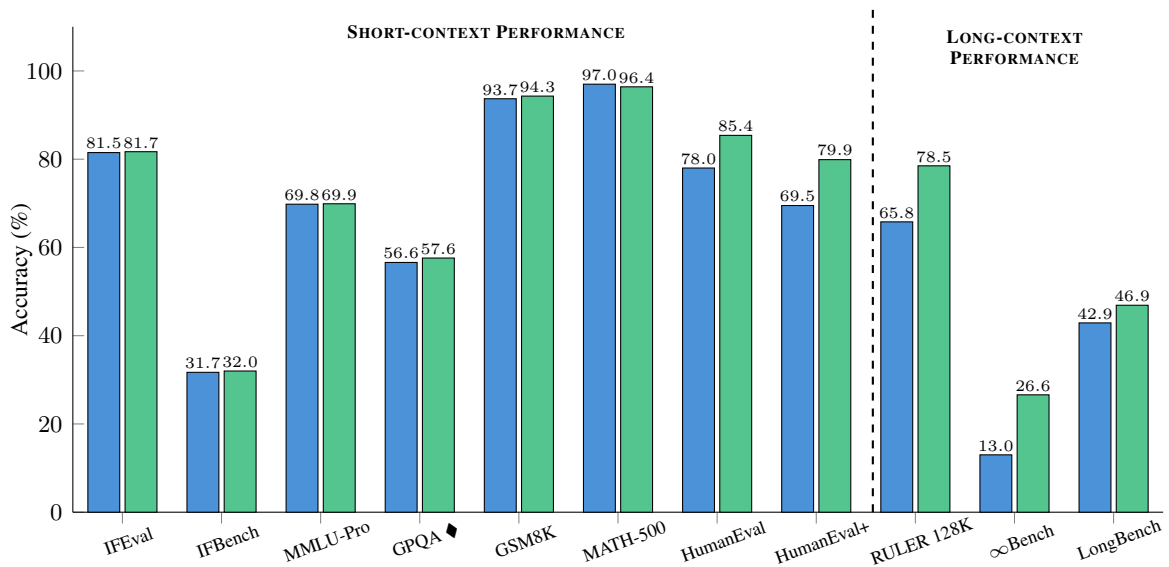
\begin{figure*}[!htbp]
    \centering
    \begin{tikzpicture}[trim axis left, trim axis right]
        \begin{axis}[
            axis lines*=left,
            ybar=2pt,
            bar width=12pt,
            width=\linewidth,
            height=8cm,
            enlarge x limits=0.05,
            ylabel={Accuracy (\%)},
            ylabel style={font=\small, yshift=-10pt},
            yticklabel style={font=\small, xshift=0pt},
            symbolic x coords={
                IFEval, IFBench, MMLU-Pro, GPQA $\blacklozenge$, GSM8K, MATH-500, HumanEval, HumanEval+,
                RULER 128K, $\infty$Bench, LongBench
            },
            xtick={
                IFEval, IFBench, MMLU-Pro, GPQA $\blacklozenge$, GSM8K, MATH-500, HumanEval, HumanEval+,
                RULER 128K, $\infty$Bench, LongBench
            },
            xticklabel style={rotate=20, anchor=center, yshift=-7.0pt, font=\scriptsize},
            legend style={
                at={(0.5,-0.3)},
                anchor=north,
                legend columns=2,
                font=\small,
                column sep=1pt,
                draw=none
            },
            ymin=0, ymax=110,
            clip=false,
            nodes near coords,
            every node near coord/.append style={
                text=black,
                font=\tiny,
                /pgf/number format/fixed,
                /pgf/number format/fixed zerofill,
                /pgf/number format/precision=1,
                anchor=center,
                yshift=3.5pt
            },
        ]

        \addplot+[fill=baselineColor, draw=black] coordinates {
            (IFEval, 81.5) (IFBench, 31.7) (MMLU-Pro, 69.8) (GPQA $\blacklozenge$, 56.6)
            (GSM8K, 93.7) (MATH-500, 97.0) (HumanEval, 78.0) (HumanEval+, 69.5)
            (RULER 128K, 65.8) ($\infty$Bench, 13.0) (LongBench, 42.9)
        };

        \addplot+[fill=ultraColor, draw=black] coordinates {
            (IFEval, 81.7) (IFBench, 32.0) (MMLU-Pro, 69.9) (GPQA $\blacklozenge$, 57.6) (GSM8K, 94.3) (MATH-500, 96.4) (HumanEval, 85.4) (HumanEval+, 79.9)
            (RULER 128K, 78.5) ($\infty$Bench, 26.6) (LongBench, 46.9)
        };

        \draw[dashed, thick] ([xshift=-16pt]axis cs:RULER 128K,0) -- ([xshift=-16pt]axis cs:RULER 128K,115);

        \node[anchor=south, font=\scriptsize\bfseries] at ([xshift=25pt]axis cs:GPQA $\blacklozenge$, 105) {\textsc{Short-context Performance}};
        \node[anchor=center, font=\scriptsize\bfseries] at (axis cs:$\infty$Bench, 105) {\makecell{\textsc{Long-context}\\\textsc{Performance}}};

        \end{axis}
    \end{tikzpicture}
    \caption{Qwen3-4B model-scaling benchmark performance.}
    \label{fig:qwen3-4b-main-results}
\end{figure*}

\end{document}